\pdfoutput=1

\documentclass[11pt]{article}

\usepackage[]{EMNLP2023}

\usepackage{times}
\usepackage{latexsym}

\usepackage[T1]{fontenc}

\usepackage[utf8]{inputenc}

\usepackage{microtype}

\usepackage{inconsolata}

\usepackage{times}
\usepackage{latexsym}
\usepackage{times}
\usepackage{comment}
\usepackage{latexsym}
\usepackage{pythonhighlight}

\usepackage{latexsym}
\usepackage{graphicx}
\usepackage{amsmath}
\usepackage{mathtools}
\usepackage{breqn}
\usepackage{amsmath}
\usepackage{url}
\usepackage{enumerate}
\definecolor{aliceblue}{rgb}{0.94, 0.97, 1.0}
\usepackage{color}
\usepackage{xcolor,colortbl}
\definecolor{green}{rgb}{0.1,0.1,0.1}
\definecolor{aliceblue}{rgb}{0.94, 0.97, 1.0}
\definecolor{Gray}{gray}{0.9}
\definecolor{GrAy}{gray}{0.96}
\definecolor{beaublue}{rgb}{0.74, 0.83, 0.9}
\definecolor{LightCyan}{rgb}{0.88,1,1}
\definecolor{inchworm}{rgb}{0.7, 0.93, 0.36}
\definecolor{babyblue}{rgb}{0.54, 0.81, 0.94}
\definecolor{blue(ncs)}{rgb}{0.0, 0.53, 0.74}
\definecolor{blond}{rgb}{0.98, 0.98, 0.82}
\usepackage{amsmath}
\usepackage{amssymb}
\usepackage{comment}
\usepackage{xspace}
\usepackage{adjustbox}
\usepackage{booktabs}
\usepackage{graphicx}
\usepackage{enumitem}
\definecolor{Gray}{gray}{0.9}
\definecolor{GrAy}{gray}{0.96}
 \definecolor{inchworm}{rgb}{0.7, 0.93, 0.36}
 \usepackage{pifont}
\usepackage{amsmath}

\newcommand*{\authorimg}[1]%
    { \raisebox{-1\baselineskip}{\includegraphics[width=\imagesize]{#1}}}
\newlength\imagesize    
\usepackage{amsmath}
\usepackage{color}
\usepackage{graphicx,array}
\newcolumntype{C}[1]{>{\centering\let\newline\\\arraybackslash\hspace{0pt}}m{#1}}
\newcolumntype{L}[1]{>{\raggedright\let\newline\\\arraybackslash\hspace{0pt}}m{#1}}
\usepackage{amsmath}
\newcolumntype{C}{ >{\vspace{-0.3cm}\centering\arraybackslash} m{4cm} }
\newcolumntype{D}{ >{\vspace{-0.1000cm}\centering\arraybackslash} m{1cm} }
\usepackage{xcoffins}
\NewCoffin\tablecoffin
\newcommand{\DrawPercentageBar}[1]{%
  \begin{tikzpicture}
    \fill[color=blue!80,opacity=0.3]   (0.0 , 0.0) rectangle (#1*3ex , 1.5ex );
    \fill[color=gray!30] (#1*3ex  , 0.0) rectangle (3.0ex, 1.5ex);
  \end{tikzpicture}%
}

\newcommand{\DrawPercentageBarNeg}[1]{%
  \begin{tikzpicture}
    \fill[color=red!40, fill opacity=0.3]   (0.0 , 0.0) rectangle (#1*3ex , 1.5ex );
    \fill[color=gray!30] (#1*3ex  , 0.0) rectangle (3.0ex, 1.5ex);
  \end{tikzpicture}%
}

\newcommand{\hlc}[2][yellow]{{%
    \colorlet{foo}{#1}%
    \sethlcolor{foo}\hl{#2}}%
}

\usepackage{xcolor,colortbl}
\definecolor{green}{rgb}{0.1,0.1,0.1}

\makeatletter
\DeclareRobustCommand\onedot{\futurelet\@let@token\@onedot}
\def\@onedot{\ifx\@let@token.\else.\null\fi\xspace}

\def\eg{\emph{e.g}\onedot} 
\def\ie{\emph{i.e}\onedot} 
 
\def\etc{\emph{etc}\onedot}

\makeatother

\usepackage{amssymb}
\usepackage{pifont}
\usepackage{booktabs}
\usepackage{nicematrix}
\usepackage{xcoffins}

  \newcolumntype{C}[1]{>{\centering\let\newline\\ \arraybackslash\hspace{0pt}}m{#1}}
\newcolumntype{L}[1]{>{\raggedright\let\newline\\ \arraybackslash\hspace{0pt}}m{#1}}

\makeatletter
\DeclareRobustCommand\onedot{\futurelet\@let@token\@onedot}
\def\@onedot{\ifx\@let@token.\else.\null\fi\xspace}

\def\eg{\emph{e.g}\onedot} 
\def\ie{\emph{i.e}\onedot} 
 
\def\etc{\emph{etc}\onedot}

\usepackage[utf8]{inputenc}

\usepackage{microtype}

\usepackage{color}
\usepackage{xcolor,colortbl}
\definecolor{aliceblue}{rgb}{0.94, 0.97, 1.0}
\definecolor{Gray}{rgb}{0.94, 0.97, 1.0}
\definecolor{beaublue}{rgb}{0.74, 0.83, 0.9}
\definecolor{LightCyan}{rgb}{0.88,1,1}
\definecolor{inchworm}{rgb}{0.7, 0.93, 0.36}
\usepackage{epstopdf}
\usepackage[utf8]{inputenc}
\newcolumntype{C}{ >{\vspace{-0.3cm}\centering\arraybackslash} m{4cm} }
\newcolumntype{D}{ >{\vspace{-0.1000cm}\centering\arraybackslash} m{1cm} }

 \newcolumntype{W}[1]{>{\centering\let\newline\\\arraybackslash\hspace{0pt}}m{#1}}
\newcolumntype{L}[1]{>{\raggedright\let\newline\\\arraybackslash\hspace{0pt}}m{#1}}
\usepackage{amssymb}
\usepackage{pifont}
\newcommand{\xmark}{\ding{56}}%
\usepackage{bbm}
\usepackage{xcolor}
\usepackage{soul}

\def\eg{\emph{e.g}\onedot} 
\def\ie{\emph{i.e}\onedot} 
 
\def\etc{\emph{etc}\onedot}

\makeatother

\DeclareRobustCommand\ie{%
  \UKUS@comma{i.e}%
}
\DeclareRobustCommand\eg{%
  \UKUS@comma{e.g}%
}
\definecolor{lightcyan}{rgb}{0.88, 1.0, 1.0}
 	\definecolor{cyan}{rgb}{0.0, 1.0, 1.0}
\usepackage{tikz}

\DeclareRobustCommand\onedot{\futurelet\@let@token\@onedot}
\def\@onedot{\ifx\@let@token.\else.\null\fi\xspace}

\usepackage{xspace}

\newcolumntype{C}{ >{\vspace{-0.3cm}\centering\arraybackslash} m{4cm} }
\newcolumntype{D}{ >{\vspace{-0.1000cm}\centering\arraybackslash} m{1cm} }

\newcolumntype{C}{ >{\vspace{-0.3cm}\centering\arraybackslash} m{4cm} }
\newcolumntype{D}{ >{\vspace{-0.1000cm}\centering\arraybackslash} m{1cm} }

\makeatletter
\DeclareRobustCommand\onedot{\futurelet\@let@token\@onedot}
\def\@onedot{\ifx\@let@token.\else.\null\fi\xspace}

\def\eg{\emph{e.g}\onedot} 
\def\ie{\emph{i.e}\onedot} 
 
\def\etc{\emph{etc}\onedot}

\usepackage[utf8]{inputenc}

\usepackage{microtype}
\usepackage{hyperref}
\usepackage{xcolor}

\usepackage{xstring}
\usepackage{color}

\usepackage{scalerel,graphicx,xparse}

\usepackage[T1]{fontenc}

\usepackage[utf8]{inputenc}

\usepackage{microtype}

\title{Women Wearing Lipstick: Measuring the Bias \\ Between an Object and Its Related Gender}

\author{Ahmed Sabir  \hspace{0.3cm} 
Llu\'{\i}s Padr\'o   \\
 Universitat Polit\`ecnica de Catalunya, TALP Research Center, Barcelona, Spain  \\ 
\tt asabir@cs.upc.edu, padro@cs.upc.edu
}

\begin{document}
\maketitle
\begin{abstract}

In this paper, we investigate the impact of objects on gender bias in image captioning systems. Our results show that only gender-specific objects have a strong gender bias (\eg \textit{women-lipstick}). In addition, we propose a visual semantic-based gender score that measures the degree of bias and can be used as a plug-in for any image captioning system. Our experiments demonstrate the utility of the gender score, since we observe that our score can measure the bias relation between a caption and its related gender; therefore, our score can be used as an additional metric to the existing  Object Gender Co-Occ approach. Code and data are publicly available at \url{https://github.com/ahmedssabir/GenderScore}.


\end{abstract}

\section{Introduction}

Visual understanding of image captioning is an important and rapidly evolving research topic \cite{karpathy2015deep, anderson2018bottom}. Recent approaches predominantly rely on Transformer \cite{huang2019attention,  cho2022fine} and the BERT based pre-trained paradigm \cite{Jacob:19} to  
learn cross-modal representation \cite{li2020oscar,zhang2021vinvl, li2022blip, li2023blip}.


While image captioning models achieved notable benchmark performance in utilizing the correlation between visual and co-occurring labels to generate an accurate image description, this often results in a gender bias that relates to a specific gender, such as confidently identifying a woman when there is a kitchen in the image. The work of \cite{zhao2017men} tackles the problem of gender bias in visual semantic role labeling by balancing the distribution. For image captioning, \citet{hendricks2018women} consider ameliorating gender bias via a balance classifier. More recently, \citet{hirota2022quantifying} measured racial and gender bias amplification in image captioning via a trainable classifier on an additional human-annotated existing caption dataset for the gender bias task \cite{zhao2021understanding}.

\begin{figure*}[ht]
\small 
\centering 

\includegraphics[width=\textwidth]{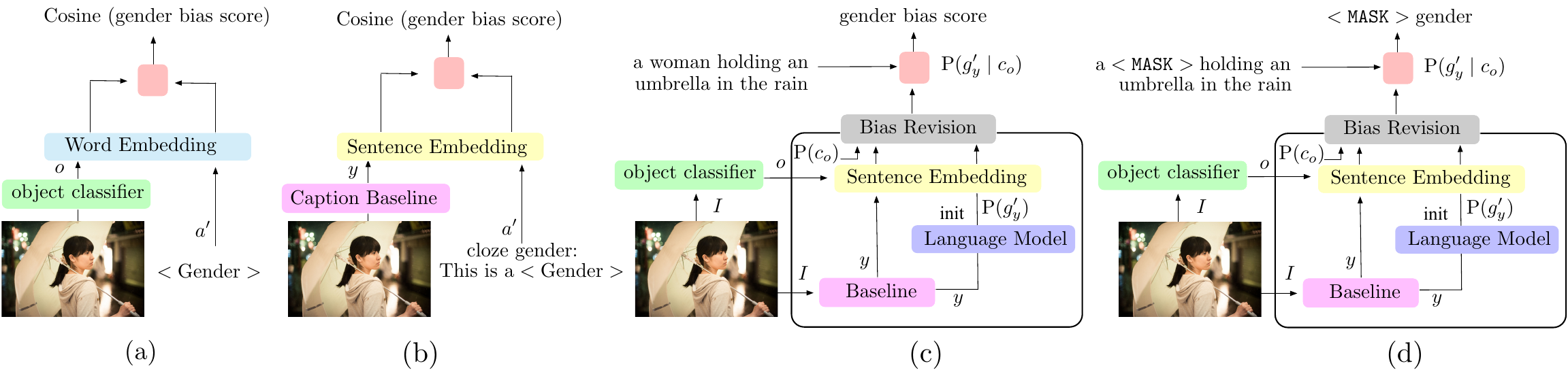} 

\vspace{-0.2cm} 

\caption{An overview of the proposed gender bias measure for image captioning. (a) word level with word embedding, (b) sentence level with semantic embedding, (c) our proposed gender score that relies on visual bias revision (with GPT-2 for initialization $\mathrm{P}\left(g^{\prime}_{y}\right)$ (common observation as initial bias without visual) and with BERT for similarity score), and (d) the proposed  < \texttt{MASK} > gender estimation (prediction), also with visual bias  revision.}

\label{fig:main_figure}
\end{figure*}

To close the full picture of gender bias in image captioning, in this work, unlike other works, we examine the problem from a visual semantic relation perspective. We therefore propose a Gender Score via human-inspired  judgment named Belief Revision \cite{blok2003probability} which can be used to (1) discover bias and (2) predict gender bias without \textit{training} or \textit{unbalancing} the dataset. We conclude our contributions are as follows: (1) we investigate the gender object relation for image captioning at the  word level (\ie object-gender) and  the sentence level with captions (\ie gender-caption); (2) we propose a Gender Score that uses gender in relation to the visual information in the image to predict the gender bias. Our Gender Score can be used as a plug-in for any out-of-the-box image captioning system. Figure \ref{fig:main_figure} shows an overview of the proposed gender bias measure for image captioning.


\section{Visual Context Information}
\label{se:data}


In this work, we investigate the relation between gender bias and the objects that are mainly used in image captioning systems, and more precisely, the widely used manually annotated image caption datasets: Flickr30K \cite{young2014image} and COCO dataset \cite{lin2014microsoft}. However, we focus mainly on the COCO dataset as the most used dataset in gender bias evaluation. COCO Captions is an unbalanced gender bias dataset with a 1:3 ratio bias towards men \cite{zhao2017men,hendricks2018women,tang2021mitigating}. The dataset contains around 120K images, and each image is annotated with five different human-written captions.

To obtain the \textit{visual context} $o$ from each image $I$, we use out-of-the-box classifiers to extract the image context information $o(I)$. Specifically, following \cite{sabir2023visual}, the objects extracted from all pre-trained models are obtained by extracting the top-3 object class/category  (excluding \textit{person} category) from each classifier after filtering out instances with (1) the cosine distance between objects (voting between classifiers of top-$k$) and (2) a low confidence score via a probability threshold $< 0.2$. We next describe each of these classifiers:

\begin{table}[h!]
\centering
\resizebox{0.99\columnwidth}{!}{
\begin{tabular}{lll|lll}

\hline 
  \multicolumn{3}{c}{GloVe Gender Dist (Biased)}  &  \multicolumn{3}{c}{GloVe Gender Dist (Balanced)}  \\ 
 \cmidrule(lr){1-3} \cmidrule(lr){4-6}
 + person & + man  & + woman & + person & + man  & + woman  \\ 
\hline 
 face & walking &     lipstick & can & police &  police\\
 hand &  cowboy&      dishwasher &police & white  & black  \\
 ballplayer  &  gorilla &      sleeping  &go &walking &white \\
 mailbox   & teddy  &      horse &walking  & black  &  car \\
  book   &  ballplayer&      bathtub & black& car  & walking \\
\hline 


\end{tabular}
}
\vspace{-0.2cm}
\caption{Most frequency count of object + gender in the COCO Captions training dataset via Cosine Distance with GloVe and balanced Gender Neutral GN-GloVe.}
\label{tb:glove}
\end{table}

\noindent{\bf ResNet-152 \cite{he2016deep}: }A residual deep network that is designed for ImageNet classification tasks, which relies heavily on batch normalization.





\noindent{\bf CLIP \cite{radford2021learning}:} (Contrastive Language-Image Pre-training) is a pre-trained model with contrastive loss where the pair of image-text needs to be distinguished from randomly selected sample pairs. CLIP uses Internet resources without human annotation on 400M pairs. 

\noindent{\bf Inception-ResNet FRCNN \cite{huang2017speed}:} is an improved variant of Faster R-CNN, with a trade-off of better accuracy and fast inference via high-level features extracted from Inception-ResNet. It is a pre-trained model that is trained on COCO categories, with 80 object categories. 


\section{Gender Object Distance}
\label{sec3:cosine}



The main component of our Gender Score (next section) is the semantic relation between the object and the gender. Therefore, in this section, we investigate the semantic correlation between the object and the gender in the general dataset (\eg wiki). More specifically,  we assume that all images contain a gender \textit{man}, \textit{woman} or gender neutral \textit{person}, and we aim to measure the Cosine Distance between the object $o$, objects with context \ie caption $y$ from the image $I$ and its related gender $a' \in\{\text {man, woman, \textit{person}}\}$. In this context, we refer to the Cosine Distance as the closest distance (\ie semantic relatedness score) between the gender-to-object via similarity measure  \eg cosine similarity. We apply our analysis to the most widely used human-annotated datasets in image captioning tasks: Flikr30K and COCO Captions datasets.

We employ several commonly used pre-trained models: (1) word-level: GloVe \cite{pennington2014glove}, fasttext \cite{bojanowski2017enriching} and word2vec \cite{mikolov2013efficient} out-of-the-box as baselines, then we utilize Gender Neutral GloVe \cite{zhao2018learning}, which balances the gender bias; (2) sentence-level: Sentence-BERT \cite{reimers2019sentence} tuned on Natural Language Inference  (NLI) \cite{conneau2017supervised}, (2b) SimCSE \cite{gao2021simcse} contrastive learning supervised by NLI, and an improved SimCSE version with additional \textbf{Info}rmation aggregation via masked language model objective InfoCSE \cite{wu2022infocse}. In particular, we measure the Cosine Distance from the gender $a'$ to the object $o$ or caption $y$ as shown in Figure \ref{fig:main_figure} (a, b) (\eg how close the gender vector of $a'$ is to the object $o$ \textit{umbrella}). Table \ref{tb:glove} shows the top most object-gender frequent count with the standard and gender-balanced GloVe, respectively.




\begin{table*}[t!]
\centering
\resizebox{0.92\textwidth}{!}{
\begin{tabular}{lcccccccccc}

\hline 

& \multicolumn{5}{c}{COCO Captions} & \multicolumn{5}{c}{Flickr30K} \\ 
 \cmidrule(lr){2-6}\cmidrule(lr){7-11} 
& \multicolumn{3}{c}{Avg: Gender Object Distance} & \multicolumn{2}{c}{Ratio} & \multicolumn{3}{c}{Avg: Gender Object Distance} & \multicolumn{2}{c}{Ratio} \\ 
 \cmidrule(lr){2-4}\cmidrule(lr){5-6} \cmidrule(lr){7-9}\cmidrule(lr){10-11}
Model & + person & + man  & + woman &  to-m & to-w & + person & + man  & + woman &  to-m & to-w  \\ 
\hline 


\hline 
Word2Vec \cite{mikolov2013efficient}  & 0.101 &  0.116 & 0.124   &  0.48  &  \cellcolor{aliceblue} 0.51 & 0.116 & 0.142 & 0.154 & 0.47 & \cellcolor{aliceblue} 0.52\\
GloVe \cite{pennington2014glove} &0.146 & 0.175 &  0.169 &  0.50 & 0.49   & 0.131 & 0.170 & 0.168 & 0.50 & 0.49\\
Fasttext \cite{bojanowski2017enriching}  & 0.180 &  0.200 & 0.191  &  \cellcolor{aliceblue}0.51 & 0.48  & 0.146  & 0.196 &  0.191 & 0.50& 0.49 \\
GN-GloVe \cite{zhao2018learning}  & 0.032 &  0.055 & 0.054 &   0.50 & 0.49 & 0.024 & 0.085 & 0.088 & 0.49 & 0.50 \\
\hline 
SBERT \cite{reimers2019sentence} & 0.124 &  0.155 & 0.128  &  \cellcolor{aliceblue}0.54 & 0.45 & 0.121 & 0.167 & 0.129 &  \cellcolor{aliceblue}0.56 & 0.43\\
SimCSE-RoBERTa \cite{gao2021simcse} & 0.194 &   0.137 & 0.093  &  \cellcolor{aliceblue}0.59 & 0.40 & 0.189 & 0.140& 0.107 &  \cellcolor{aliceblue}0.56 & 0.43 \\
InfoCSE-RoBERTa \cite{wu2022infocse} & 0.199 &   0.222 & 0.211  &  \cellcolor{aliceblue}0.51 & 0.48 & 0.228 & 0.265 & 0.241  &  \cellcolor{aliceblue}0.52 & 0.47 \\
\hline 
\end{tabular}
}
\vspace{-0.2cm}


\caption{Result of Average Cosine Distance between gender and object in the COCO Captions and Flikr30K training datasets. The ratio is the gender \hlc[aliceblue]{bias rate} \textbf{to}wards men/women. The results show there is a slight bias toward men. GloVe and GN-GloVe (balanced) show identical results on COCO Captions, which indicate that not all objects have a strong bias toward a specific gender. In particular, regarding non-biased objects,  both models exhibit a low/similar bias ratio \eg \textit{bicycle}-\textit{gender} (GloVe: m=0.31 | w=0.27, ratio=0.53) and (GN-GloVe: m=0.15 | w=0.13, ratio=0.53).}

\label{tb:caption_result}
\end{table*}

\begin{table*}[t!]
\centering
\resizebox{0.74\textwidth}{!}{
\begin{tabular}{lcccccccc}

\hline 

& \multicolumn{3}{c}{Avg: Gender Score} & \multicolumn{3}{c}{Bias Ratio} &  \multicolumn{2}{c}{Leakage } \\ 
 \cmidrule(lr){2-4}\cmidrule(lr){5-7}\cmidrule(lr){8-9}
Model & + person & + man  & + woman & m & w & to-m & m & w \\ 
\hline 

Human &  &    &    &  &   &   & 930 & 291 \\
\addlinespace[0.1cm]

\multicolumn{9}{l}{\textbf{Gender Object Distance}}  \\ 
Transformer \cite{Ashish:17}    & 0.075  & 0.062  &  0.033 &  0.82 & 0.44  & 0.65 &  0.85 & 1.40  \\
AoANet \cite{huang2019attention}  & 0.079 &  0.061 & 0.047   & 0.77 & 0.59  & 0.56  &  0.82 & 1.29 \\
Vilbert \cite{lu202012}  & 0.089 &  0.070 & 0.056   & 0.78 & 0.62  &  0.55  & 0.75 &  1.06 \\
OSCAR \cite{li2020oscar}  & 0.076 & 0.054 & 0.044   &  0.71 & 0.57  & 0.58  & 0.93 & 1.40 \\
BLIP \cite{li2022blip} & 0.068 &  0.050 & 0.040  & 0.73 & 0.58  &  0.55  & 0.83 & 1.32  \\
TraCLIPS-Reward \cite{cho2022fine} & 0.069 &  0.053 & 0.040  & 0.76 & 0.57 &   0.56  & 0.82 & 1.30 \\
BLIP-2 \cite{li2023blip} & 0.073 &  0.049 & 0.043   & 0.67 & 0.58  & 0.53  & 0.73 & 1.22 \\

\hline 
\addlinespace[0.1cm]
\multicolumn{9}{l}{\textbf{Gender Score}}  \\ 
Transformer \cite{Ashish:17}  & 0.201 & 0.194 &  0.187 &  0.96 & 0.93 & \cellcolor{aliceblue}0.50 &  0.85 & 1.40 \\
AoANet \cite{huang2019attention}  & 0.217 &  0.210 & 0.201   & 0.96 & 0.92  & \cellcolor{aliceblue}0.51  &  0.82 & 1.29 \\
Vilbert \cite{lu202012}  & 0.225 &  0.217 & 0.207   & 0.96 & 0.92  & \cellcolor{aliceblue}0.51  & 0.75 &  1.06  \\
OSCAR \cite{li2020oscar}  & 0.196 &  0.188 & 0.181   & 0.95 & 0.92  & \cellcolor{aliceblue}0.50  & 0.93 & 1.40  \\
BLIP \cite{li2022blip} & 0.226 &  0.220 & 0.213   & 0.97 & 0.94  & \cellcolor{aliceblue}0.50 & 0.83 & 1.32 \\
TraCLIPS-Reward \cite{cho2022fine} & 0.198 &  0.194 & 0.185  & 0.97 & 0.93 &    \cellcolor{aliceblue}0.51  & 0.82 & 1.30 \\
BLIP-2 \cite{li2023blip} & 0.224 & 0.215 &  0.210 &   0.95 & 0.93   & \cellcolor{aliceblue}0.50  &  0.73 & 1.22 \\
\hline 
\end{tabular}

}

\vspace{-0.2cm}

\caption{Result of the Average Cosine Distance and Gender Score, between gender and object information, on the Karpathy test split. Our score \hlc[aliceblue]{balances} the bias better than direct Cosine Distance, primarily because not all objects exhibit strong gender bias. The leakage is a comparison between human-annotated caption and the classifier output that uses the gender-related object to influence the final gender prediction,  such as associating women with food.}




\label{tb:caption_test}
\end{table*}

\noindent{\bf Gender Bias Ratio:} we follow the work of \cite{zhao2017men} to calculate the gender  bias ratio towards men as:
\begin{equation}
\small 
\text{bias}_{\text{to-$m$}} = \frac{\operatorname{s}(\text{obj, $m$})}{\operatorname{s}(\text{obj, $m$})+\operatorname{s}(\text{obj, $w$})}
\end{equation}



where $m$  and $w$ refer to the gender in the image, and the $s$ is our proposed gender-to-object semantic relation bias score. In our case, we also use the score to compute the ratio to gender neutral  \textit{person}:

\begin{equation}
\small 
\text{Ratio}_{\text{to-n}} = \frac{\operatorname{s}(\text{obj, $m$/$w$})}{\operatorname{s}(\text{obj, \text{\textit{person}}})}
\end{equation}

\section{Gender Score}
\label{se:gender-score}



In this section, we describe the proposed \textbf{G}ender \textbf{S}core that estimates gender based on its semantic relation  with the visual information extracted from the image. \citet{sabir2022belief} proposed a caption re-ranking method that leverages visual semantic context.  This approach utilized Belief Revision \cite{blok2003probability} to convert the similarity \ie Cosine Distance (Section \ref{sec3:cosine})  into a probability measure.

\noindent{\textbf{Belief Revision.}} To obtain likelihood bias revisions based on similarity scores (\ie gender, object), we need three parameters: (1) \textbf{Hypothesis (g)}: caption $y$ with the associated gender $a \in\{\text {man, woman}\}$, (2) \textbf{Informativeness (c)}: image object information $ o$ confidence and (3) \textbf{Similarities}: the degree of relatedness between  object and gender $sim(y,o)$.

\begin{equation}
\small
\text{GS}_{a}(y) = \frac{1}{|\mathcal{D}|} \sum_{(y, o) \in \mathcal{D}} \operatorname{P}\left(g_{y} \mid c_{o}\right) = \operatorname{P}(g_{y})^{\alpha}
\label{ch2: sim-to-prob-1}
\end{equation}



\noindent where $\operatorname{P}(g_{y})$ is the \textit{hypothesis} probability of $y$, $\mathcal{D}$ is the predicted captions  with the gender $a$, and  $\operatorname{P}(c_{o})$ is the probability of the evidence that causes hypothesis probability revision \ie visual bias 
revision via  object context $o$ from the image $I$, $o(I)$:





\noindent{\textbf{Hypothesis}:} $\operatorname{P}(g_{y}) \quad (\small{\text{caption}} \, y \, \small{\text{with the gender}}\, a)$ 

\noindent{\textbf{Informativeness}:} $1-\operatorname{P}(c_{o}) \quad  (\small{\text{object context}} \, o$) 




\noindent{\textbf{Similarities}:}  $\alpha=\left[\frac{1-\operatorname{sim}(y, o)}{1+\operatorname{sim}(y, o)}\right]^{1-\operatorname{P}(c_{o})} (\small{\text{{visual bias}}})$ 





\noindent{The visual} context $\operatorname{P}(c_{o})$ will revise the caption with the associated gender $\operatorname{P}(g_{y})$  (\ie gender bias) if there is a semantic relation between them $sim(y,o)$. We discuss each component next:

 \begin{table*}[t!]
\centering
\resizebox{0.9\textwidth}{!}{
\begin{tabular}{lcccccccc}

\hline 


& \multicolumn{4}{c}{Bias Ratio Toward Men} & \multicolumn{4}{c}{Bias Ratio Toward Women}  \\ 
 \cmidrule(lr){2-5}\cmidrule(lr){6-9}
Model & skateboard &  kitchen  & motorcycle & baseball &  skateboard &  kitchen  & motorcycle & baseball \\ 

\hline 

\addlinespace[0.1cm]

\multicolumn{9}{l}{Object Gender Co-Occ  \cite{zhao2017men} } \\ 
\addlinespace[0.1cm]
Transformer \cite{Ashish:17}    &  0.96  \DrawPercentageBar{0.96}  & 0.50  \DrawPercentageBar{0.50}  & 0.83  \DrawPercentageBar{0.41}  & 0.75 \DrawPercentageBar{0.75} &  0.05 \DrawPercentageBarNeg{0.05}  & 0.50 \DrawPercentageBarNeg{0.50} &    0.16 \DrawPercentageBarNeg{0.16}   &  0.25 \DrawPercentageBarNeg{0.25}  \\
AoANet \cite{huang2019attention}  & 0.97  \DrawPercentageBar{0.97} &  0.51 \DrawPercentageBar{0.51} & 0.85 \DrawPercentageBar{0.85}  & 0.81 \DrawPercentageBar{0.81}  &  0.02 \DrawPercentageBarNeg{0.02}   & 0.48 \DrawPercentageBarNeg{0.48}   & 0.14 \DrawPercentageBarNeg{0.14}  &  0.18 \DrawPercentageBarNeg{0.18} \\
Vilbert \cite{lu202012}  &  0.96 \DrawPercentageBar{0.96} &   0.47 \DrawPercentageBar{0.47} &  0.84 \DrawPercentageBar{0.84}   &  0.66 \DrawPercentageBar{0.66} &  0.03 \DrawPercentageBarNeg{0.03}   & 0.52 \DrawPercentageBarNeg{0.52}   &  0.15 \DrawPercentageBarNeg{0.15}  &  0.33 \DrawPercentageBarNeg{0.33} \\
OSCAR \cite{li2020oscar}  & 0.97 \DrawPercentageBar{0.97} &  0.58 \DrawPercentageBar{0.58} & 0.82 \DrawPercentageBar{0.82}  & 0.90 \DrawPercentageBar{0.90}& 0.02 \DrawPercentageBarNeg{0.02}  & 0.41 \DrawPercentageBarNeg{0.41}  &  0.18 \DrawPercentageBarNeg{0.18} & 0.09 \DrawPercentageBarNeg{0.09}\\

BLIP \cite{li2022blip} &  0.96 \DrawPercentageBar{0.96} &   0.52 \DrawPercentageBar{0.52} & 0.88 \DrawPercentageBar{0.88}   &  0.97 \DrawPercentageBar{0.97} &  0.03 \DrawPercentageBarNeg{0.03}  &  0.47 \DrawPercentageBarNeg{0.47}    & 0.11 \DrawPercentageBarNeg{0.11}  &  0.02 \DrawPercentageBarNeg{0.02}  \\



TraCLIPS-Reward \cite{cho2022fine} &  0.89 \DrawPercentageBar{0.89} &   0.48 \DrawPercentageBar{0.48} &  0.93 \DrawPercentageBar{0.93}   &  0.50 \DrawPercentageBar{0.50} &  0.10 \DrawPercentageBarNeg{0.10}  &  0.51 \DrawPercentageBarNeg{0.51}    & 0.06 \DrawPercentageBarNeg{0.06   }  &  0.50 \DrawPercentageBarNeg{0.50}  \\

BLIP-2 \cite{li2023blip} &  0.94 \DrawPercentageBar{0.94} &   0.57 \DrawPercentageBar{0.57} & 0.88 \DrawPercentageBar{0.88}   &  0.90 \DrawPercentageBar{0.90} &  0.05 \DrawPercentageBarNeg{0.05}  &  0.42 \DrawPercentageBarNeg{0.42}    & 0.11 \DrawPercentageBarNeg{0.11}  &   0.10   \DrawPercentageBarNeg{0}  \\

\hline 
\addlinespace[0.1cm]
\multicolumn{9}{l}{Gender Score}  \\
\addlinespace[0.1cm]
Transformer \cite{Ashish:17}    & 0.96  \DrawPercentageBar{0.96}  &  0.51  \DrawPercentageBar{0.51}  &  0.83  \DrawPercentageBar{0.83}  & 0.61 \DrawPercentageBar{0.61}  & 0.03 \DrawPercentageBarNeg{0.03}   & 0.48 \DrawPercentageBarNeg{0.48} &    0.16 \DrawPercentageBarNeg{0.16} & 0.38 \DrawPercentageBarNeg{0.38} \\
AoANet \cite{huang2019attention}  & 0.97  \DrawPercentageBar{0.97}  &  0.46  \DrawPercentageBar{0.46}  & 0.84 \DrawPercentageBar{0.84}    & 0.82 \DrawPercentageBar{0.82}  & 0.02 \DrawPercentageBarNeg{0.02}  &  0.53 \DrawPercentageBarNeg{0.53}   & 0.15 \DrawPercentageBarNeg{0.15} & 0.17 \DrawPercentageBarNeg{0.17}  \\
Vilbert \cite{lu202012}  &  0.96 \DrawPercentageBar{0.96}  &  0.53 \DrawPercentageBar{0.53} & 0.84 \DrawPercentageBar{0.84}   & 0.65 \DrawPercentageBar{0.65} &   0.03 \DrawPercentageBarNeg{0.03} & 0.46 \DrawPercentageBarNeg{0.46}  & 0.15 \DrawPercentageBarNeg{0.15}  &  0.34 \DrawPercentageBarNeg{0.34}  \\
OSCAR \cite{li2020oscar}  & 0.98 \DrawPercentageBar{0.98} &  0.42 \DrawPercentageBar{0.42} &  0.78  \DrawPercentageBar{0.78}   & 0.83 \DrawPercentageBar{0.83} & 0.01 \DrawPercentageBar{0.01}  &  0.57 \DrawPercentageBarNeg{0.57}  & 0.21  \DrawPercentageBarNeg{0.21}  &   0.16 \DrawPercentageBarNeg{0.16}  \\

BLIP \cite{li2022blip} &  0.96 \DrawPercentageBar{0.96} &  0.50 \DrawPercentageBar{0.50} & 0.86 \DrawPercentageBar{0.86}   &  0.98 \DrawPercentageBar{0.98} &  0.03 \DrawPercentageBarNeg{0.03}  &  0.49 \DrawPercentageBarNeg{0.49}  & 0.13 \DrawPercentageBarNeg{0.13} & 0.01 \DrawPercentageBarNeg{0.01}  \\



TraCLIPS-Reward \cite{cho2022fine} &  0.88 \DrawPercentageBar{0.88} &   0.43 \DrawPercentageBar{0.43} & 0.92 \DrawPercentageBar{0.92}   &  0.50 \DrawPercentageBar{0.50} &  0.11 \DrawPercentageBarNeg{0.11}  &  0.56 \DrawPercentageBarNeg{0.56}    & 0.07 \DrawPercentageBarNeg{0}  &  0.49 \DrawPercentageBarNeg{0.49}  \\

BLIP-2 \cite{li2023blip} &  0.93 \DrawPercentageBar{0.93} &   0.56 \DrawPercentageBar{0.56} & 0.82 \DrawPercentageBar{0.82}   & 0.89  \DrawPercentageBar{0.89} &  0.06 \DrawPercentageBarNeg{0.06}  &  0.43 \DrawPercentageBarNeg{0.43}    & 0.17 \DrawPercentageBarNeg{0.17}  &  0.10 \DrawPercentageBarNeg{0.10}  \\

\hline

\end{tabular}
}

\vspace{-0.2cm}


\caption{Example of the most common gender bias objects in COCO Captions, Karpathy test split. The result shows that our score (bias ratio) aligns closely with the existing Object Gender Co-Occ approach when applied to the most gender-biased objects toward men. Note that TraCLIPS-Reward (CLIPS+CIDEr) inherits biases from RL-CLIPS, resulting in distinct gender predictions and generates caption w/o a specific gender \ie \textit{person},  \textit{baseball player},  etc. }

\label{tb:caption-example}
\end{table*}



\noindent{\textbf{Hypothesis initial bias}:} In visual-based belief revision, one of the conditions is to start with an initial hypothesis and then revise it using visual context and a similarity score. Therefore, we initialize the caption hypothesis  $\operatorname{P}(g_{y})$ with a common observation $\operatorname{P}(g'_{y})$, such as a Language Model (LM) (we consider this as an initial bias without visual). We employ (GPT-2) \cite{radford2019language} with mean token probability since it achieves better results.

\noindent{\bf Informativeness of bias information:} As the visual context probability $\operatorname{P}(c_{o})$ approaches 1 and in consequence is less informative (very frequent objects have no discriminative power since they may co-occur with any gender) $1-\operatorname{P}(c_{o})$  approaches zero, causing $\alpha$ to get closer to 1, and thus, a smaller revision of $\operatorname{P}(g'_{y})$. Therefore, as we described in the Visual Context Information Section \ref{se:data}, we leverage a threshold and semantic filter visual context dataset from  ResNet, Inception-ResNet v2 based Faster R-CNN object detector, and CLIP to extract top-$k$  textual visual context information from the image. The extracted object is used to measure the gender-to-object bias direct  relation.








\noindent{\bf Relatedness between hypothesis and bias information:} Likelihood revision occurs if there is a close correlation between the hypothesis and the new information. As the $sim(y,o)$ (gender, object), gets closer to 1 (higher relatedness) $\alpha$ gets closer to 0, and thus hypothesis probability is revised (\ie gender bias) and raised closer to 1. Therefore, the initial hypothesis will be revised or backed off to 1 (no bias) based on the relatedness score. In our case, we employ Sentence-BERT to compute the Cosine Distance (Section \ref{sec3:cosine}) by using object $o$ as context for the caption $y$ with associated gender $a$.

\section{Experiments}





\noindent{\bf Caption model.} We examine seven of the most recent Transformer  state-of-the-art caption models: Transformer \cite{Ashish:17} with bottom-up top-down features \cite{anderson2018bottom}, AoANet \cite{huang2019attention}, Vilbert  \cite{lu202012}, OSCAR  \cite{li2020oscar}, BLIP \cite{li2022blip}, Transformer with Reinforcement Learning (RL) CLIPS+CIDEr as image+text similarity Reward (TraCLIPS-Reward) \cite{cho2022fine} and Large LM$_{\text{2.7B}}$ based BLIP-2 \cite{li2023blip}. Note that for a fair comparison with other pre-trained models, OSCAR uses a cross-entropy evaluation score rather than the RL-based CIDEr optimization score.






\noindent{\bf Data.} Our gender bias score is performed on the (82783 $\times$ 5 human annotations) COCO Captions dataset. For baselines (testing), the score is used to evaluate gender-to-object bias on the standard 5K Karpathy test split images \cite{karpathy2015deep} (GT is the average of five human bias ratios).

Our experiments apply visual semantics between the object or object with context \ie caption and its related gender information to predict object-to-gender related bias. The proposed visual-gender scores are (1): Cosine Distance, which uses the similarity score to try to estimate the proper gender as shown in Table \ref{tb:caption_result}; (2) Gender Score, which carries out Belief Revision (visual bias likelihood revision) to revise the hypothesis initial bias  via similarity \eg Cosine Distance (gender, object). For our baseline, we adopt the Object Gender Co-Occ metric \cite{zhao2017men} for the image captioning task.

In this work, we hypothesize that every image has a gender $\in\{\text {man, woman}\}$ or gender neutral  (\ie \textit{person}), and our rationale is that each model suffers from \textit{Right for the Wrong Reasons} \cite{hendricks2018women} or leakage \cite{wang2019balanced} (see Table \ref{tb:caption_test}), such as associating all kitchens with women. Therefore, we want to explore all the cases and let the proposed distance/score decide which gender (\ie bias) is in the image based on a visual bias. In particular, inspired by the cloze probability last word completion task \cite{gonzalez2007methods}, we generate two identical sentences but with a different gender, and then we compute the likelihood revisions between the sentence-gender and the caption using the object probability. Table~\ref{tb:caption_result} shows that GloVe and GN-GloVe (balanced) have identical results on COCO Captions dataset, which indicate that not all objects have a strong bias toward a specific gender. In addition, Table~\ref{tb:caption_test} shows that our score balances the bias of the Cosine Distance and demonstrates, on average, that not all objects have a strong gender bias. Also, our approach detects strong specific object-to-gender bias and has a similar result to the existing Object Gender Co-Occ method on the most biased object toward men, as shown in Table~\ref{tb:caption-example}. TraCLIPS-Reward inherits biases from RL-CLIPS and thus generates caption w/o a specific gender (\eg  \textit{person}, \textit{guy}, etc). Therefore, we adopt the combined CLIPS+CIDEr Rewards, which suffer less gender prediction error.




\begin{figure}[t!]
\small


\resizebox{\columnwidth}{!}{
 \begin{tabular}{W{3.3cm}  L{5cm}}

   
  



\includegraphics[width=\linewidth, height = 2.5 cm]{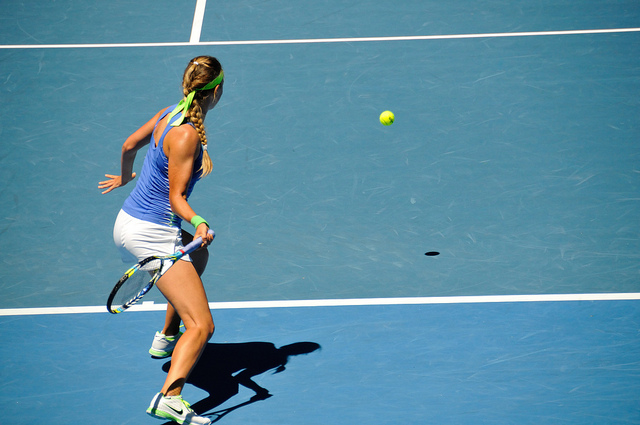} & \textbf{Visual:} tennis ball  \newline   \textbf{Caption:}   a  < \texttt{MASK} >  hitting \newline a tennis   ball on    a tennis court  \newline 
   
   
    \textbf{Gender Object Distance:}  \newline  man  0.44 \DrawPercentageBar{0.44} \space woman \textbf{0.46} \DrawPercentageBarNeg{0.46}
  \newline \textbf{Gender Score:} 
  
  man  0.45 \DrawPercentageBar{0.45} \space woman  0.45 \DrawPercentageBarNeg{0.45} \\
  
        \arrayrulecolor{gray}\hline

       \vspace{-0.2cm} 
   \includegraphics[width=\linewidth, height = 2.5 cm]{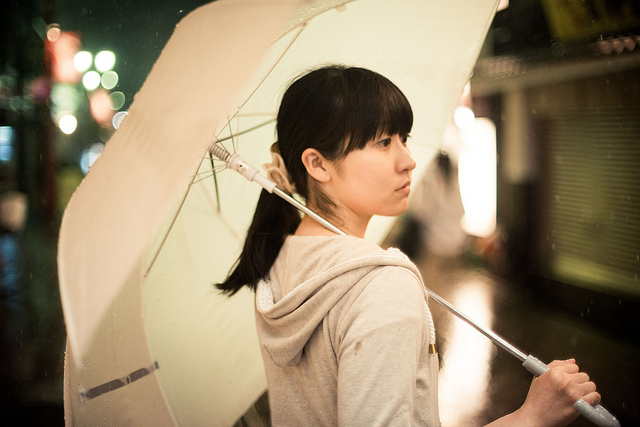} & \textbf{Visual:}  umbrella
  \newline   \textbf{Caption:}    a  < \texttt{MASK} > holding \newline  an umbrella  in the rain  \newline


    \textbf{Gender Object Distance:}  \newline  man  0.20 \DrawPercentageBar{0.20} \space woman  0.20 \DrawPercentageBarNeg{0.20}
  \newline \textbf{Gender Score:} 
  
  man  0.12 \DrawPercentageBar{0.12 } \space woman  \textbf{0.13} \DrawPercentageBarNeg{0.13} \\

        \arrayrulecolor{gray}\hline 
          
     \arrayrulecolor{gray}\hline 
       \vspace{-0.2cm} 
  \includegraphics[width=\linewidth, height = 2.5 cm]{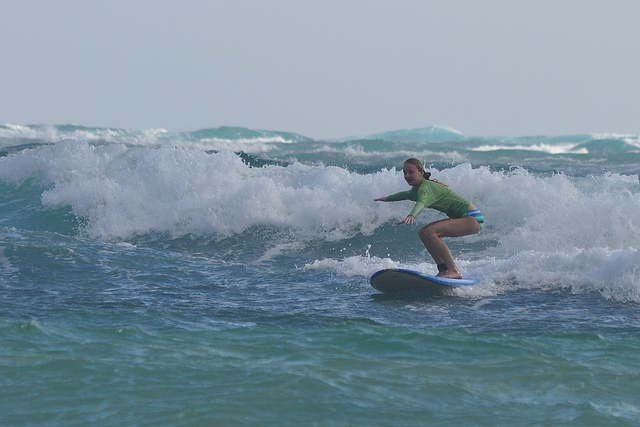} & \textbf{Visual:} paddle  \newline   \textbf{Caption:}  a < \texttt{MASK} > riding a wave \newline on top  of a surfboard  \newline

    \textbf{Gender Object Distance:}  \newline  man \textcolor{red}{0.16} \DrawPercentageBar{0.16} \space woman  \textcolor{black}{0.11} \DrawPercentageBarNeg{0.11}
  \newline \textbf{Gender Score:} 
  
 man \textcolor{red}{0.33}  \DrawPercentageBar{0.33} \space woman  0.30 \DrawPercentageBarNeg{0.30} \\
    \end{tabular}
}
 \vspace{-0.3cm}

\caption{Examples of Gender Score Estimation and Cosine Distance. The result shows that (Top) the score balances the bias (as men and women have a similar bias for the sport \textit{tennis}), (Middle) a slight bias object \textit{umbrella} toward women, and (Bottom) men strong object bias relation (\textit{paddle, surfboard}), the model adjusts the women bias while preserving the object gender score.}





\label{tb-fig:gender-est}
\end{figure}

\begin{table}[t!]\centering
\resizebox{0.74\columnwidth}{!}{
\begin{tabular}{lcccc}
\toprule 
& \multicolumn{2}{c}{Gender} & \multicolumn{2}{c}{Bias Ratio}  \\ 
 \cmidrule(lr){2-3}\cmidrule(lr){4-5}
Model & man & woman &  to-m &  to-w\\
\midrule 
\addlinespace[0.1cm]
\multicolumn{5}{l}{Object Gender Co-Occ  \cite{zhao2017men}}	\\ 

\addlinespace[0.1cm]
Transformer  &     792    	&  408	 &  0.66 	&  0.34  \\
AoANet   &     770    	&  368	 &  0.67 	&  0.32   \\
Vilbert  &    702     	&  311	 &  0.69  	&  0.30   \\
OSCAR  &     845    	&  409	 &  0.67	&  0.32   \\
BLIP & 775 &  385 & 0.66   & 0.33  \\
TraCLIPS-Reward  & 769 &  381 & 0.66   & 0.33 \\
BLIP-2 & 695 &  356 & 0.66    & 0.33 \\
\hline 
\multicolumn{5}{l}{Gender Score (Gender Score Estimation) }	\\ 
\addlinespace[0.1cm]
Transformer  &     616   	&  217	 &  \cellcolor{aliceblue}0.73 	&  0.26   \\
AoANet   &         527	&  213	 &  \cellcolor{aliceblue}0.71	&  0.28   \\
Vilbert  &    526     	&  161	 &  \cellcolor{aliceblue}0.76	&  0.23   \\
OSCAR  &     630   	&  237	 &  {\cellcolor{aliceblue}}0.72	&  0.27   \\
BLIP & 554 &  240 & \cellcolor{aliceblue}0.69   & 0.30  \\
TraCLIPS-Reward  & 537 &  251 & \cellcolor{aliceblue}0.68   & 0.31 \\
BLIP-2 & 498 &  239 & \cellcolor{aliceblue}0.67   & 0.32 \\
\bottomrule 

\end{tabular}
}
\vspace{-0.2cm}


\caption{Comparison results between Object Gender Co-Occ (predicted gender-object results) and our gender score estimation on Karpathy test split. The proposed score measures gender bias more \hlc[aliceblue]{accurately}, particularly when there is a strong object to gender bias relation.}

\label{tb:benh-mark}

\end{table}

\noindent{\bf{Gender Score Estimation.}} In this experiment, we < \texttt{MASK} > the gender and use the object with context to predict the gender as shown in Figure \ref{tb-fig:gender-est}. The idea is to measure the amplified gender-to-object bias in pre-trained models. Table  \ref{tb:benh-mark} shows that the fill-in gender score has a more  bias towards men results than object-gender pair counting.  The rationale is that the model can estimate the strong gender object bias as shown in Figure \ref{tb-fig:gender-est}, including the false positive and leakage cases by the classifier.


\noindent{\bf{Discussion.}} Our approach measures the amplified gender bias more accurately than the Object Gender Co-Occ \cite{zhao2017men} in the following two scenarios: (1) where the gender is not obvious in the image and is misclassified by the caption baseline, and (2) when there is leakage by the classifier. In addition, unlike the Object Gender Co-Occ, our model balances the gender to object bias and only measures a strong object to gender bias relation as shown in Figure \ref{tb-fig:gender-est}. For instance, the word “hitting” in a generated caption (as a stand-alone without context) is associated with a bias toward men more than women, and it will influence the final gender-to-caption bias score. However, our gender score balances the unwanted bias and only measures the pronounced gender to object bias relation. 

\section{Conclusion}

In this work, we investigate the relation between objects and gender bias in image captioning systems. Our results show that not all objects exhibit a strong gender bias and only in special cases does an object have a strong gender bias. We also propose a Gender Score that can be used as an additional metric to the existing Object-Gender Co-Occ method.


\section*{Limitations}


The accuracy of our Gender Score heavily relies on the semantic relation (\ie Cosine Distance) between gender and object, a high or low degree of similarity score between them can influence the final bias score negatively.  Also, our model relies on a large pre-trained model(s), which inherently encapsulate their own latent biases that might impact the visual bias revision behavior in gender-to-object bias scenarios. Specifically, in cases when there are multiple strong bias contexts (\ie non-objects context) with high similarity scores toward a particular gender present within the caption. This can imbalance the final gender-to-object score, leading to errors in gender prediction and bias estimation. In addition, the false positive object context information extracted by the visual classifier will result in inaccurate bias estimation.

\section*{Ethical Considerations}


We rely upon an existing range of well-known publicly available caption datasets crawled from the web and annotated by humans that assume a binary conceptualization of gender. Therefore, it is important to acknowledge that within the scope of this work, we are treating gender as strictly binary (\ie man and woman) oversimplifies a complex and multifaceted aspect of human identity. Gender is better understood as a spectrum, with many variations beyond just two categories, and should be addressed in future work. Since all models are being trained on these datasets, we anticipate all models contain other biases (racial, cultural, \etc). For example, an observation in Table  \ref{tb:glove}, when we remove gender bias, we notice the emergence of another bias, such as racial bias. For example, vectors representing \textit{Black} person or women are closer together than those representing other colors, like \textit{white}. Moreover, there is another form of bias that has received limited attention in the literature, the propagation of gender and racial bias via RL (\eg RL-CLIPS). For instance, in Figure \ref{fig:main_figure} the model associates gender with race, as “asian woman”.
\bibliography{anthology,custom}

\begin{thebibliography}{35}
\expandafter\ifx\csname natexlab\endcsname\relax\def\natexlab#1{#1}\fi

\bibitem[{Anderson et~al.(2018)Anderson, He, Buehler, Teney, Johnson, Gould, and Zhang}]{anderson2018bottom}
Peter Anderson, Xiaodong He, Chris Buehler, Damien Teney, Mark Johnson, Stephen Gould, and Lei Zhang. 2018.
\newblock Bottom-up and top-down attention for image captioning and visual question answering.
\newblock In \emph{CVPR}.

\bibitem[{Blok et~al.(2003)Blok, Medin, and Osherson}]{blok2003probability}
Sergey Blok, Douglas Medin, and Daniel Osherson. 2003.
\newblock Probability from similarity.
\newblock In \emph{AAAI}.

\bibitem[{Bojanowski et~al.(2017)Bojanowski, Grave, Joulin, and Mikolov}]{bojanowski2017enriching}
Piotr Bojanowski, Edouard Grave, Armand Joulin, and Tomas Mikolov. 2017.
\newblock Enriching word vectors with subword information.
\newblock \emph{TACL}.

\bibitem[{Cho et~al.(2022)Cho, Yoon, Kale, Dernoncourt, Bui, and Bansal}]{cho2022fine}
Jaemin Cho, Seunghyun Yoon, Ajinkya Kale, Franck Dernoncourt, Trung Bui, and Mohit Bansal. 2022.
\newblock Fine-grained image captioning with clip reward.
\newblock \emph{arXiv preprint arXiv:2205.13115}.

\bibitem[{Conneau et~al.(2017)Conneau, Kiela, Schwenk, Barrault, and Bordes}]{conneau2017supervised}
Alexis Conneau, Douwe Kiela, Holger Schwenk, Loic Barrault, and Antoine Bordes. 2017.
\newblock Supervised learning of universal sentence representations from natural language inference data.
\newblock \emph{arXiv preprint arXiv:1705.02364}.

\bibitem[{Devlin et~al.(2019)Devlin, Chang, Lee, and Toutanova}]{Jacob:19}
Jacob Devlin, Ming-Wei Chang, Kenton Lee, and Kristina Toutanova. 2019.
\newblock Bert: Pre-training of deep bidirectional transformers for language understanding.
\newblock In \emph{NAACL-HLT}.

\bibitem[{Gao et~al.(2021)Gao, Yao, and Chen}]{gao2021simcse}
Tianyu Gao, Xingcheng Yao, and Danqi Chen. 2021.
\newblock {SimCSE}: Simg of sentence embeddings.
\newblock \emph{arXiv preprint arXiv:2104.08821}.

\bibitem[{Gonzalez-Marquez(2007)}]{gonzalez2007methods}
Monica Gonzalez-Marquez. 2007.
\newblock \emph{Methods in cognitive linguistics}, volume~18.
\newblock John Benjamins Publishing.

\bibitem[{He et~al.(2016)He, Zhang, Ren, and Sun}]{he2016deep}
Kaiming He, Xiangyu Zhang, Shaoqing Ren, and Jian Sun. 2016.
\newblock Deep residual learning for image recognition.
\newblock In \emph{CVPR}.

\bibitem[{Hendricks et~al.(2018)Hendricks, Burns, Saenko, Darrell, and Rohrbach}]{hendricks2018women}
Lisa~Anne Hendricks, Kaylee Burns, Kate Saenko, Trevor Darrell, and Anna Rohrbach. 2018.
\newblock Women also snowboard: Overcoming bias in captioning models.
\newblock In \emph{ECCV}.

\bibitem[{Hirota et~al.(2022)Hirota, Nakashima, and Garcia}]{hirota2022quantifying}
Yusuke Hirota, Yuta Nakashima, and Noa Garcia. 2022.
\newblock Quantifying societal bias amplification in image captioning.
\newblock In \emph{CVPR}.

\bibitem[{Huang et~al.(2017)Huang, Rathod, Sun, Zhu, Korattikara, Fathi, Fischer, Wojna, Song, Guadarrama et~al.}]{huang2017speed}
Jonathan Huang, Vivek Rathod, Chen Sun, Menglong Zhu, Anoop Korattikara, Alireza Fathi, Ian Fischer, Zbigniew Wojna, Yang Song, Sergio Guadarrama, et~al. 2017.
\newblock Speed/accuracy trade-offs for modern convolutional object detectors.
\newblock In \emph{CVPR}.

\bibitem[{Huang et~al.(2019)Huang, Wang, Chen, and Wei}]{huang2019attention}
Lun Huang, Wenmin Wang, Jie Chen, and Xiao-Yong Wei. 2019.
\newblock Attention on attention for image captioning.
\newblock In \emph{CVPR}.

\bibitem[{Karpathy and Fei-Fei(2015)}]{karpathy2015deep}
Andrej Karpathy and Li~Fei-Fei. 2015.
\newblock Deep visual-semantic alignments for generating image descriptions.
\newblock In \emph{CVPR}.

\bibitem[{Li et~al.(2023)Li, Li, Savarese, and Hoi}]{li2023blip}
Junnan Li, Dongxu Li, Silvio Savarese, and Steven Hoi. 2023.
\newblock Blip-2: Bootstrapping language-image pre-training with frozen image encoders and large language models.
\newblock \emph{arXiv preprint arXiv:2301.12597}.

\bibitem[{Li et~al.(2022)Li, Li, Xiong, and Hoi}]{li2022blip}
Junnan Li, Dongxu Li, Caiming Xiong, and Steven Hoi. 2022.
\newblock Blip: Bootstrapping language-image pre-training for unified vision-language understanding and generation.
\newblock \emph{arXiv preprint arXiv:2201.12086}.

\bibitem[{Li et~al.(2020)Li, Yin, Li, Zhang, Hu, Zhang, Wang, Hu, Dong, Wei et~al.}]{li2020oscar}
Xiujun Li, Xi~Yin, Chunyuan Li, Pengchuan Zhang, Xiaowei Hu, Lei Zhang, Lijuan Wang, Houdong Hu, Li~Dong, Furu Wei, et~al. 2020.
\newblock Oscar: Object-semantics aligned pre-training for vision-language tasks.
\newblock In \emph{ECCV}.

\bibitem[{Lin et~al.(2014)Lin, Maire, Belongie, Hays, Perona, Ramanan, Doll{\'a}r, and Zitnick}]{lin2014microsoft}
Tsung-Yi Lin, Michael Maire, Serge Belongie, James Hays, Pietro Perona, Deva Ramanan, Piotr Doll{\'a}r, and C~Lawrence Zitnick. 2014.
\newblock Microsoft coco: Common objects in context.
\newblock In \emph{ECCV}.

\bibitem[{Lu et~al.(2020)Lu, Goswami, Rohrbach, Parikh, and Lee}]{lu202012}
Jiasen Lu, Vedanuj Goswami, Marcus Rohrbach, Devi Parikh, and Stefan Lee. 2020.
\newblock 12-in-1: Multi-task vision and language representation learning.
\newblock In \emph{CVPR}.

\bibitem[{Mikolov et~al.(2013)Mikolov, Chen, Corrado, and Dean}]{mikolov2013efficient}
Tomas Mikolov, Kai Chen, Greg Corrado, and Jeffrey Dean. 2013.
\newblock Efficient estimation of word representations in vector space.
\newblock \emph{arXiv preprint arXiv:1301.3781}.

\bibitem[{Pennington et~al.(2014)Pennington, Socher, and Manning}]{pennington2014glove}
Jeffrey Pennington, Richard Socher, and Christopher~D Manning. 2014.
\newblock Glove: Global vectors for word representation.
\newblock In \emph{EMNLP}.

\bibitem[{Radford et~al.(2021)Radford, Kim, Hallacy, Ramesh, Goh, Agarwal, Sastry, Askell, Mishkin, Clark et~al.}]{radford2021learning}
Alec Radford, Jong~Wook Kim, Chris Hallacy, Aditya Ramesh, Gabriel Goh, Sandhini Agarwal, Girish Sastry, Amanda Askell, Pamela Mishkin, Jack Clark, et~al. 2021.
\newblock Learning transferable visual models from natural language supervision.
\newblock \emph{arXiv preprint arXiv:2103.00020}.

\bibitem[{Radford et~al.(2019)Radford, Wu, Child, Luan, Amodei, and Sutskever}]{radford2019language}
Alec Radford, Jeffrey Wu, Rewon Child, David Luan, Dario Amodei, and Ilya Sutskever. 2019.
\newblock Language models are unsupervised multitask learners.
\newblock \emph{OpenAI blog}.

\bibitem[{Reimers and Gurevych(2019)}]{reimers2019sentence}
Nils Reimers and Iryna Gurevych. 2019.
\newblock Sentence-bert: Sentence embeddings using siamese bert-networks.
\newblock \emph{arXiv preprint arXiv:1908.10084}.

\bibitem[{Sabir et~al.(2022)Sabir, Moreno-Noguer, Madhyastha, and Padr{\'o}}]{sabir2022belief}
Ahmed Sabir, Francesc Moreno-Noguer, Pranava Madhyastha, and Llu{\'\i}s Padr{\'o}. 2022.
\newblock Belief revision based caption re-ranker with visual semantic information.
\newblock \emph{arXiv preprint arXiv:2209.08163}.

\bibitem[{Sabir et~al.(2023)Sabir, Moreno-Noguer, and Padr{\'o}}]{sabir2023visual}
Ahmed Sabir, Francesc Moreno-Noguer, and Llu{\'\i}s Padr{\'o}. 2023.
\newblock Visual semantic relatedness dataset for image captioning.
\newblock In \emph{CVPRW}.

\bibitem[{Tang et~al.(2021)Tang, Du, Li, Liu, Zou, and Hu}]{tang2021mitigating}
Ruixiang Tang, Mengnan Du, Yuening Li, Zirui Liu, Na~Zou, and Xia Hu. 2021.
\newblock Mitigating gender bias in captioning systems.
\newblock In \emph{Proceedings of the Web Conference 2021}.

\bibitem[{Vaswani et~al.(2017)Vaswani, Shazeer, Parmar, Uszkoreit, Jones, Gomez, Kaiser, and Polosukhin}]{Ashish:17}
Ashish Vaswani, Noam Shazeer, Niki Parmar, Jakob Uszkoreit, Llion Jones, Aidan~N Gomez, {\L}ukasz Kaiser, and Illia Polosukhin. 2017.
\newblock Attention is all you need.
\newblock In \emph{NeurIPS}.

\bibitem[{Wang et~al.(2019)Wang, Zhao, Yatskar, Chang, and Ordonez}]{wang2019balanced}
Tianlu Wang, Jieyu Zhao, Mark Yatskar, Kai-Wei Chang, and Vicente Ordonez. 2019.
\newblock Balanced datasets are not enough: Estimating and mitigating gender bias in deep image representations.
\newblock In \emph{CVPR}.

\bibitem[{Wu et~al.(2022)Wu, Gao, Lin, Han, Wang, and Hu}]{wu2022infocse}
Xing Wu, Chaochen Gao, Zijia Lin, Jizhong Han, Zhongyuan Wang, and Songlin Hu. 2022.
\newblock Infocse: Information-aggregated contrastive learning of sentence embeddings.
\newblock \emph{arXiv preprint arXiv:2210.06432}.

\bibitem[{Young et~al.(2014)Young, Lai, Hodosh, and Hockenmaier}]{young2014image}
Peter Young, Alice Lai, Micah Hodosh, and Julia Hockenmaier. 2014.
\newblock From image descriptions to visual denotations: New similarity metrics for semantic inference over event descriptions.
\newblock \emph{TACL}.

\bibitem[{Zhang et~al.(2021)Zhang, Li, Hu, Yang, Zhang, Wang, Choi, and Gao}]{zhang2021vinvl}
Pengchuan Zhang, Xiujun Li, Xiaowei Hu, Jianwei Yang, Lei Zhang, Lijuan Wang, Yejin Choi, and Jianfeng Gao. 2021.
\newblock Vinvl: Making visual representations matter in vision-language models.
\newblock In \emph{CVPR}.

\bibitem[{Zhao et~al.(2021)Zhao, Wang, and Russakovsky}]{zhao2021understanding}
Dora Zhao, Angelina Wang, and Olga Russakovsky. 2021.
\newblock Understanding and evaluating racial biases in image captioning.
\newblock In \emph{CVPR}.

\bibitem[{Zhao et~al.(2017)Zhao, Wang, Yatskar, Ordonez, and Chang}]{zhao2017men}
Jieyu Zhao, Tianlu Wang, Mark Yatskar, Vicente Ordonez, and Kai-Wei Chang. 2017.
\newblock Men also like shopping: Reducing gender bias amplification using corpus-level constraints.
\newblock \emph{arXiv preprint arXiv:1707.09457}.

\bibitem[{Zhao et~al.(2018)Zhao, Zhou, Li, Wang, and Chang}]{zhao2018learning}
Jieyu Zhao, Yichao Zhou, Zeyu Li, Wei Wang, and Kai-Wei Chang. 2018.
\newblock Learning gender-neutral word embeddings.
\newblock \emph{arXiv preprint arXiv:1809.01496}.

\end{thebibliography}
\bibliographystyle{acl_natbib}

 \section{Appendix}

 \section*{Additional Information}

 \noindent{\textbf{If the goal is to evaluate gender bias why using pre-trained models.}}
We believe this is representative of real-world scenarios where we can measure the gender bias from real data and not from a closed domain as in the human-annotated COCO Captions dataset. Moreover, in contrast to recent previous work \cite{hirota2022quantifying}, which heavily depends on a costly human annotation dataset \cite{zhao2021understanding}, limiting the scalability across various domains, our Gender Score is not bounded by any dataset and can be adapted for any task, including the additionally annotated existing caption dataset for the gender bias analysis.

\begin{figure}[t!]
\small


\resizebox{\columnwidth}{!}{
 \begin{tabular}{W{3.3cm}  L{5cm}}

        \includegraphics[width=\linewidth, height = 2.5 cm]{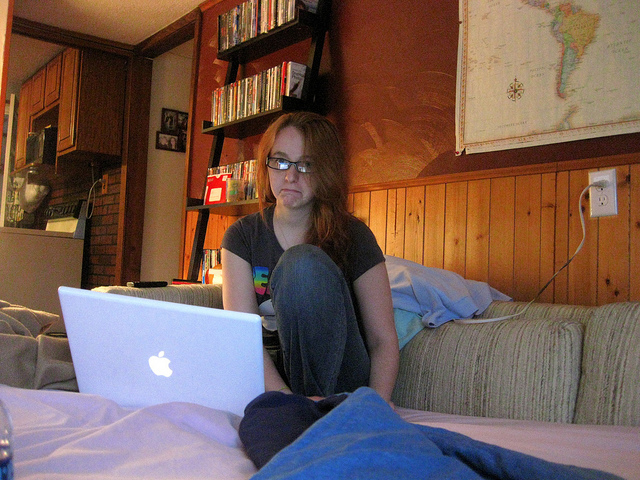} & \textbf{Visual:} laptop  \newline   \textbf{Caption:} a   < \texttt{MASK} > sitting on \newline a  couch with two laptops  \newline   \textbf{BERT} \texttt{[MASK]}: woman  
   \newline

    \textbf{Gender Object Distance:}  \newline  man  \textcolor{red}{0.43} \DrawPercentageBar{0.43} \space woman  0.42 \DrawPercentageBarNeg{0.42}
  \newline \textbf{Gender Score:} 
  
  man  0.25 \DrawPercentageBar{0.25} \space woman  0.25 \DrawPercentageBarNeg{0.25} \\  
          \arrayrulecolor{gray}\hline 
       \vspace{-0.2cm} 
   \includegraphics[width=\linewidth, height = 2.5 cm]{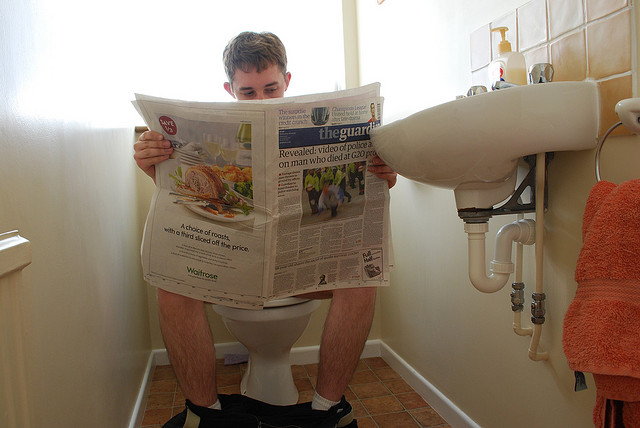} & \textbf{Visual:} toilet seat  \newline   \textbf{Caption:}    a  < \texttt{MASK} > reading \newline  a paper in a bathroom \newline 
   \textbf{BERT} \texttt{[MASK]}: man

    \textbf{Gender Object Distance:}  \newline  man  \textbf{0.42} \DrawPercentageBar{0.42} \space woman  0.40  \DrawPercentageBarNeg{0.40}
  \newline \textbf{Gender Score:} 
  
  man  0.15 \DrawPercentageBar{0.15} \space woman  \textcolor{red}{0.16} \DrawPercentageBarNeg{0.16} \\
    \arrayrulecolor{gray}\hline 
       \vspace{-0.2cm} 
   \includegraphics[width=\linewidth, height = 2.5 cm]{COCO_val2014_000000100977.jpg} & \textbf{Visual:} zebra \textcolor{red}{\xmark} \newline   \textbf{Caption:} a  < \texttt{MASK} > feeding \newline  a giraffe through a fence \newline \textbf{BERT} \texttt{[MASK]}:  dog  \newline

    \textbf{Gender Object Distance:}  \newline  man  0.23 \DrawPercentageBar{0.23} \space woman  \textbf{0.26} \DrawPercentageBarNeg{0.26}
  \newline \textbf{Gender Score:} 
  
  man  0.30 \DrawPercentageBar{0.30} \space woman  \textbf{0.32} \DrawPercentageBarNeg{0.32} \\
  
  \arrayrulecolor{gray}\hline 
       \vspace{-0.2cm} 
   \includegraphics[width=\linewidth, height = 2.5 cm]{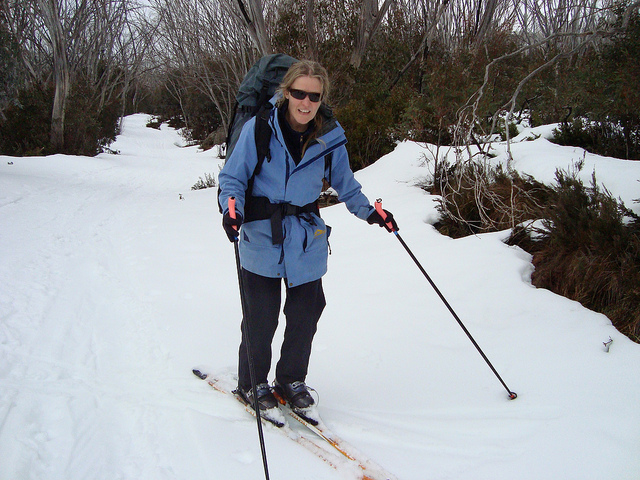} & \textbf{Visual:} walking stick \newline   \textbf{Caption:}  a  < \texttt{MASK} > riding skis \newline down   a snow covered slop  \newline 
   
  \textbf{BERT} \texttt{[MASK]}:  man  \newline 
   
    \textbf{Cosine Distance:}  \newline  man  0 \DrawPercentageBar{0} \space woman  \textcolor{black}{0} \DrawPercentageBarNeg{0.016}
  \newline \textbf{Gender Score:} 
  
  man  \textbf{0.11} \DrawPercentageBar{0.11} \space woman  0.09 \DrawPercentageBarNeg{0.10} \\
  
  \arrayrulecolor{gray}\hline 
       \vspace{-0.2cm} 
   \includegraphics[width=\linewidth, height = 2.5 cm]{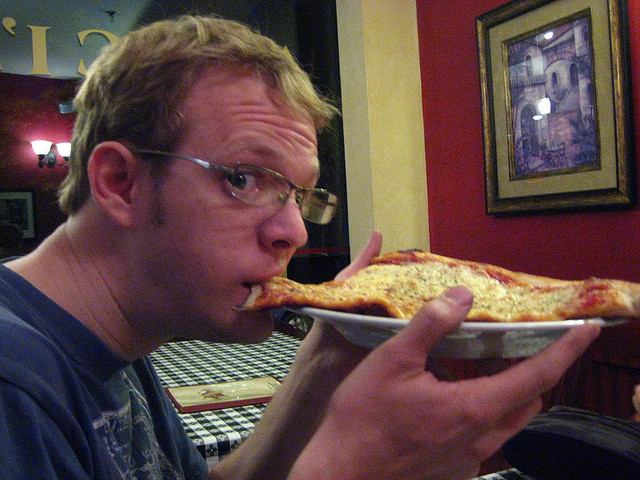} & \textbf{Visual:}    pizza 
  \newline   \textbf{Caption:}   a  < \texttt{MASK} >  eating \newline a slice   of pizza   \newline 
   
   
 \textbf{BERT} \texttt{[MASK]}:  man  \newline

    \textbf{Gender Object Distance:}  \newline  man  \textbf{0.62} \DrawPercentageBar{0.62} \space woman  0.57 \DrawPercentageBarNeg{0.57}
  \newline \textbf{Gender Score:} 
  
  man  \textbf{0.39} \DrawPercentageBar{0.39} \space woman  0.36 \DrawPercentageBarNeg{0.36} \\
  
    
  \arrayrulecolor{gray}\hline 
       \vspace{-0.2cm} 
   \includegraphics[width=\linewidth, height = 2.5 cm]{COCO_val2014_000000020254.jpg} & \textbf{Visual:}  umbrella  \newline   \textbf{Caption:}  a  < \texttt{MASK} > holding \newline a hot dog and a bottle \textcolor{red}{\xmark}
   
  \textbf{BERT} \texttt{[MASK]}:  woman  \newline

    \textbf{Gender Object Distance:}  \newline  man  0.15 \DrawPercentageBar{0.15} \space woman  \textcolor{red}{0.20} \DrawPercentageBarNeg{0.20}
  \newline \textbf{Gender Score:} 
  
  man   0.23 \DrawPercentageBar{0.23} \space woman   \textcolor{red}{0.24} \DrawPercentageBarNeg{0.24} \\

  \arrayrulecolor{gray}\hline 
       \vspace{-0.2cm} 
   \includegraphics[width=\linewidth, height = 2.5 cm]{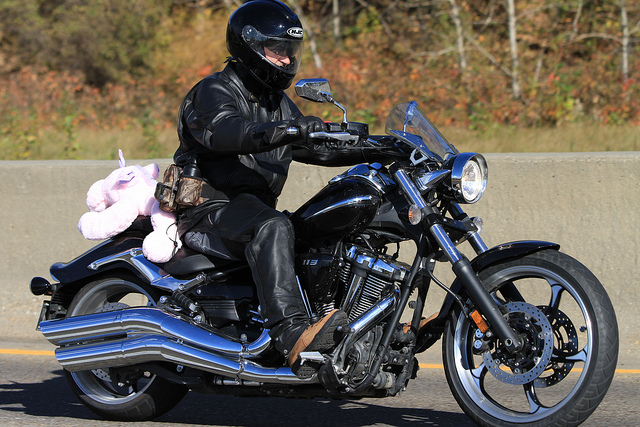} & \textbf{Visual:} motor scooter 
\newline   \textbf{Caption:} a  < \texttt{MASK} > riding a \newline motorcycle  on a road    \newline 
   
    \textbf{BERT} \texttt{[MASK]}:  man  \newline

    \textbf{Gender Object Distance:}  \newline  man  \textbf{0.54} \DrawPercentageBar{0.54} \space woman  \textcolor{black}{0.50} \DrawPercentageBarNeg{0.50}
  \newline \textbf{Gender Score:}

  man  \textbf{0.43} \DrawPercentageBar{0.43} \space woman  0.40  \DrawPercentageBarNeg{0.40} \\ 
   \arrayrulecolor{gray}\hline 
       \vspace{-0.2cm} 
   \includegraphics[width=\linewidth, height = 2.5 cm]{COCO_val2014_000000175864.jpg} & \textbf{Visual:} tennis ball   \newline   \textbf{Caption:}  a  < \texttt{MASK} > sitting on \newline a bench    holding a tennis racket   \newline

 \textbf{BERT} \texttt{[MASK]}:  man  \newline

    \textbf{Gender Object Distance:}  \newline  man  0.45 \DrawPercentageBar{0.45} \space woman  \textcolor{red}{0.51} \DrawPercentageBarNeg{0.51}
  \newline \textbf{Gender Score:} 
  
  man  0.47 \DrawPercentageBar{0.47} \space woman   \textcolor{red}{0.50} \DrawPercentageBarNeg{0.50} \\
 
    \end{tabular}
}
 \vspace{-0.2cm}


\caption{Examples of Gender Score Estimation and Cosine Distance. In the \textit{Laptop} example, our score balances the amplified bias. However, our score is less successful in balancing the bias in \textit{tennis} instance, as there are strong  biases  of non-object contexts associated with women (\textit{sitting} and \textit{holding}).  Our baseline is BERT as Mask Language Modeling to predict the \texttt{Mask} using only the word surrounding context without any visual information. For example, for the same \textit{Laptop} instance, the word \textit{couch} has a strong bias toward \textit{women}, which influences the gender output and is similar to gender to \textit{motorcycle} prediction which relies on a \textit{men} bias.}

\label{tb-fig:3}

\end{figure}

\begin{figure}[t!]
\small


\resizebox{\columnwidth}{!}{
 \begin{tabular}{W{3.3cm}  L{5cm}}

        \includegraphics[width=\linewidth, height = 2.5 cm]{COCO_val2014_000000257870.jpg} & \textbf{Visual:}  hotdog 
 \newline   \textbf{Caption:} a  < \texttt{MASK} > eating a hot dog  \newline 
   \textbf{BERT} \texttt{[MASK]}:  dog  \newline

    \textbf{Gender Object Distance:}  \newline  man  \textbf{0.44} \DrawPercentageBar{0.42} \space woman  0.42 \DrawPercentageBarNeg{0.42}
  \newline \textbf{Gender Score:} 
  
  man  \textbf{0.42} \DrawPercentageBar{0.42} \space woman  0.40  \DrawPercentageBarNeg{0.40} \\  
          \arrayrulecolor{gray}\hline 
       \vspace{-0.2cm} 
   \includegraphics[width=\linewidth, height = 2.5 cm]{COCO_val2014_000000045550.jpg} & \textbf{Visual:} spaghetti squash \textcolor{red}{\xmark} \newline   \textbf{Caption:} a  < \texttt{MASK} > holding \newline  a sandwich and french fries \newline 
  \textbf{BERT} \texttt{[MASK]}:  waiter (man) \newline

    \textbf{Gender Object Distance:}  \newline  man  0.17 \DrawPercentageBar{0.17} \space woman \textcolor{red}{0.24}  \DrawPercentageBarNeg{0.24}
  \newline \textbf{Gender Score:}

  man  0.24 \DrawPercentageBar{0.24} \space woman  \textcolor{red}{0.39} \DrawPercentageBarNeg{0.39} \\
    \arrayrulecolor{gray}\hline 
       \vspace{-0.2cm} 
       
   \includegraphics[width=\linewidth, height = 2.5 cm]{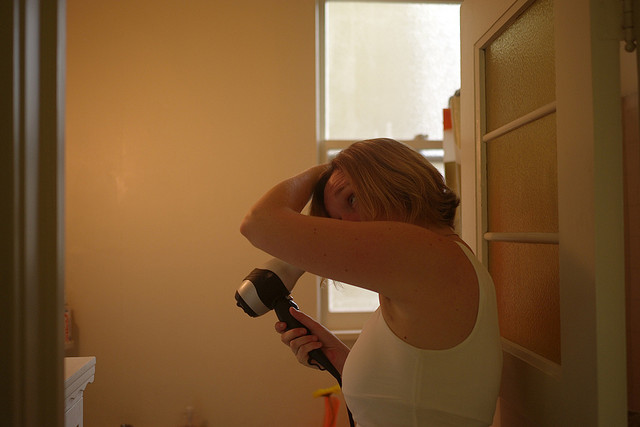} & \textbf{Visual:} hair spray   \newline   \textbf{Caption:} a  < \texttt{MASK} > holding \newline a baby    in a chair \textcolor{red}{\xmark} \newline 
 \textbf{BERT} \texttt{[MASK]}:  woman  \newline 
   
    \textbf{Cosine Distance:}  \newline  man 0.24 \DrawPercentageBar{0.24} \space woman  \textbf{0.39} \DrawPercentageBarNeg{0.39}
  \newline \textbf{Gender Score:}

  man  0.30 \DrawPercentageBar{0.30} \space woman  \textbf{0.32} \DrawPercentageBarNeg{0.32} \\
  
  \arrayrulecolor{gray}\hline 
       \vspace{-0.2cm} 
   \includegraphics[width=\linewidth, height = 2.5 cm]{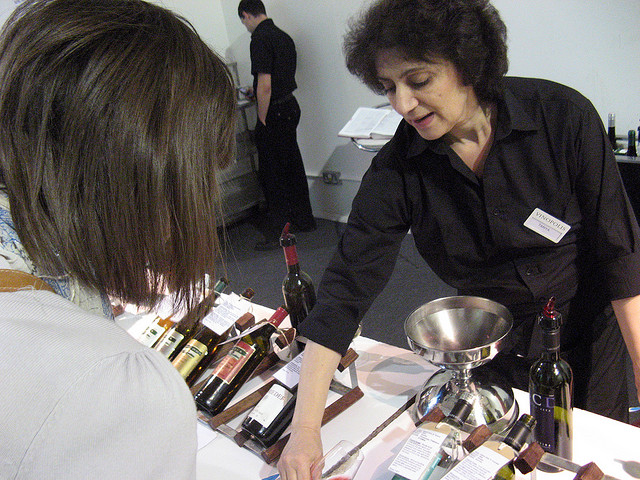} & \textbf{Visual:}  wine bottle 
\newline   \textbf{Caption:}  a  < \texttt{MASK} > sitting at \newline a table preparing food \textcolor{red}{\xmark} \newline 
   
\textbf{BERT} \texttt{[MASK]}:  woman  \newline

    \textbf{Gender Object Distance:}  \newline  man  0 \DrawPercentageBar{0} \space woman  \textcolor{black}{0} \DrawPercentageBarNeg{0.016}
  \newline \textbf{Gender Score:} 
  
  man  0.05 \DrawPercentageBar{0.05} \space woman  \textbf{0.06} \DrawPercentageBarNeg{0.06} \\
  
  \arrayrulecolor{gray}\hline 
       \vspace{-0.2cm} 
   \includegraphics[width=\linewidth, height = 2.5 cm]{COCO_val2014_000000201637.jpg} & \textbf{Visual:}  umbrella
  \newline   \textbf{Caption:}    a  < \texttt{MASK} > holding \newline  an umbrella  in the rain  \newline 
   


\textbf{BERT} \texttt{[MASK]}:  man  \newline

    \textbf{Gender Object Distance:}  \newline  man  0.20 \DrawPercentageBar{0.20} \space woman  0.20 \DrawPercentageBarNeg{0.20}
  \newline \textbf{Gender Score:} 
  
  man  0.12 \DrawPercentageBar{0.12 } \space woman  \textbf{0.13} \DrawPercentageBarNeg{0.13} \\
  
    
  \arrayrulecolor{gray}\hline 
       \vspace{-0.2cm} 
   \includegraphics[width=\linewidth, height = 2.5 cm]{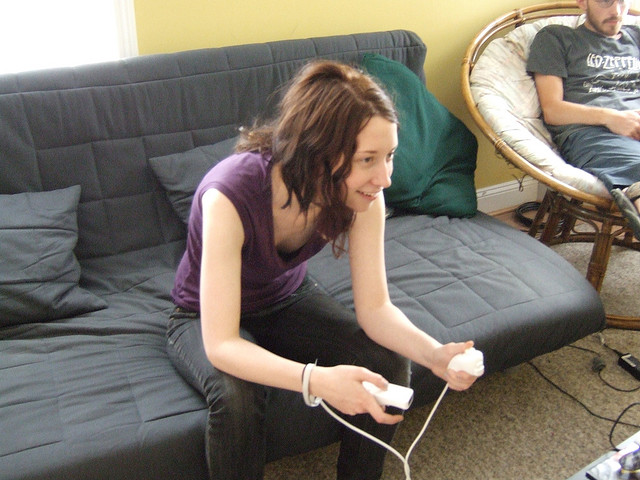} & \textbf{Visual:}  remote control   \newline   \textbf{Caption:}  a  < \texttt{MASK} > playing  \newline a video game in a living room 
   
 \textbf{BERT} \texttt{[MASK]}:  man  \newline

    \textbf{Gender Object Distance:}  \newline  man  \textcolor{red}{0.24} \DrawPercentageBar{0.24} \space woman  \textcolor{black}{0.14} \DrawPercentageBarNeg{0.14}
  \newline \textbf{Gender Score:} 
  
  man   \textcolor{red}{0.38} \DrawPercentageBar{0.38} \space woman   0.31 \DrawPercentageBarNeg{0.31} \\

  \arrayrulecolor{gray}\hline 
       \vspace{-0.2cm} 
   \includegraphics[width=\linewidth, height = 2.5 cm]{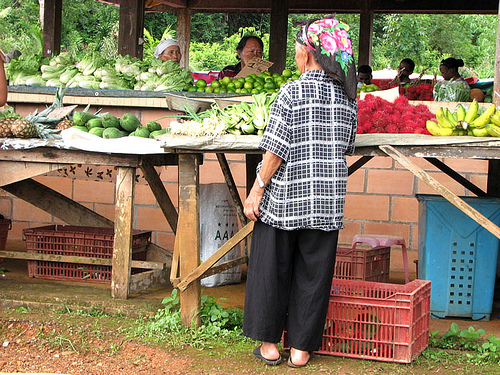} & \textbf{Visual:} shopping basket 
\newline   \textbf{Caption:}   a  < \texttt{MASK} > standing in \newline  front of a fruit stand  \newline 
   
   \textbf{BERT} \texttt{[MASK]}:  man  \newline

    \textbf{Gender Object Distance:}  \newline  man  0.34 \DrawPercentageBar{0.34} \space woman  \textbf{0.38} \DrawPercentageBarNeg{0.38}
  \newline \textbf{Gender Score:}

  man  0.46 \DrawPercentageBar{0.46} \space woman  \textbf{0.48}  \DrawPercentageBarNeg{0.48} \\ 
   \arrayrulecolor{gray}\hline 
       \vspace{-0.2cm} 
   \includegraphics[width=\linewidth, height = 2.5 cm]{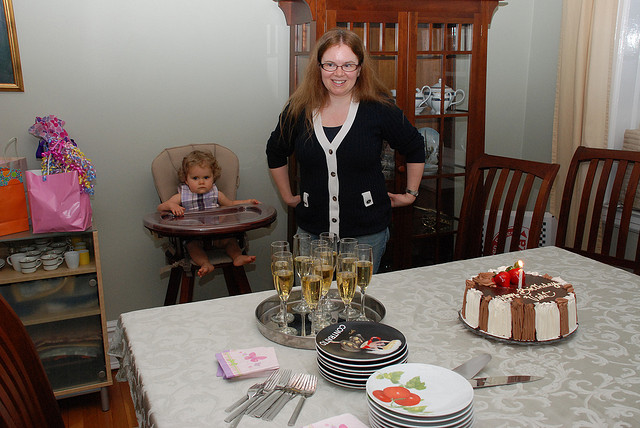} & \textbf{Visual:} dining table    \newline   \textbf{Caption:}   a  < \texttt{MASK} > and a baby at \newline a table   with a cake  \newline

 \textbf{BERT} \texttt{[MASK]}: mother (woman)  \newline 


    \textbf{Gender Object Distance:}  \newline  man  0.26 \DrawPercentageBar{0.26} \space woman  \textbf{0.40} \DrawPercentageBarNeg{0.40}
  \newline \textbf{Gender Score:} 
  
  man  0.14 \DrawPercentageBar{0.14} \space woman  \textbf{0.16} \DrawPercentageBarNeg{0.16} \\
 
    \end{tabular}
}
 \vspace{-0.2cm}

\caption{Examples of Gender Score Estimation and Cosine Distance. Note that in some cases when there is a no Cosine Distance $\leq$ 0 (Gender Object Distance) between the gender and the caption, the model will rely on general observation language model GPT-2 \ie initial bias without the visual context as the main gender bias, as shown in the \textit{wine bottle-woman} example and in Figure \ref{tb-fig:3}  the ski \textit{walking stick-woman} instance.}


\label{tb-fig:4}


\end{figure}

Table \ref{tb:benh-additional} shows a comparison result between our Gender Score and BERT$_{\text{L}}$ /GPT-2$_{\text{L}}$ as baselines using only the word surrounding context without any visual bias. The result shows that our gender estimation model is similar to BERT in some cases as shown in Figures \ref{tb-fig:3}|\ref{tb-fig:4}, which are only influenced by a strong bias. However,  when the generated caption is more/less diverse, \ie due to less/more context to predict the gender, BERT \texttt{Mask} LM struggles with predicting the correct biased gender. In contrast to BERT, our approach relies on additional context extracted from the image and uses a language model and similarity score to measure the degree of gender-based bias prediction.

\begin{table}[t!]\centering
\resizebox{0.75\columnwidth}{!}{
\begin{tabular}{lcccc}
\toprule 
& \multicolumn{2}{c}{Gender} & \multicolumn{2}{c}{Bias Ratio}  \\ 
 \cmidrule(lr){2-3}\cmidrule(lr){4-5}
Model & man & woman &  to-m &  to-w\\
\midrule 

\multicolumn{4}{l}{ Baseline - GPT-2 Score (initial bias)}	\\ 
\addlinespace[0.1cm]
Transformer  & 654  &     161    	& 0.80	 &  0.19	  \\
AoANet   &        686   	& 175	 &  0.79	&  0.20    \\
Vilbert   &        450   	& 148	 &  0.75	&  0.24    \\
OSCAR   &      645     	& 215	 &  0.75	&  0.25   \\
BLIP & 545 &  220  &   0.71  & 0.28  \\
TraCLIPS-Reward  &606 &  205 & 0.74   & 0.25 \\
BLIP-2 & 522 &  218 & 0.70   & 0.29 \\
\hline

\multicolumn{4}{l}{ Baseline - BERT MLM (no visual)}	\\ 
\addlinespace[0.1cm]
Transformer  & 524 &     187    	& \cellcolor{blond}0.73	 &  \cellcolor{blond}0.26	  \\
AoANet   &        531    	& 172	 &  \cellcolor{blond}0.75	&  \cellcolor{blond}0.24    \\
Vilbert   &        517    	& 158	 &  \cellcolor{blond}0.76	&  \cellcolor{blond}0.23    \\
OSCAR   &        576    	& 205	 & \cellcolor{blond} 0.73	& \cellcolor{blond}0.26   \\
BLIP & 541 &  195  &  0.73  & 0.26  \\
TraCLIPS-Reward  &559 &  180 & 0.75   & 0.24 \\
BLIP-2 & 536&  181 & 0.74   & 0.25 \\
\hline 
\addlinespace[0.1cm]
\multicolumn{4}{l}{ Object Gender Co-Occ  \cite{zhao2017men}}	\\ 
\addlinespace[0.1cm]
Transformer  &     792    	&  408	 &  0.66	&  0.34   \\
AoANet   &     770    	&  368	 &  0.67	&  0.32   \\
Vilbert  &    702     	&  311	 &  0.69	&  0.30   \\
OSCAR  &     845    	&  409	 &  0.67	&  0.32   \\
BLIP & 775 &  385 & 0.66   & 0.33  \\
TraCLIPS-Reward  & 769 &  381 & 0.66  & 0.33 \\
BLIP-2 & 695 &  356 & 0.66   & 0.33 \\

\hline 
\multicolumn{4}{l}{Gender Score (Gender estimation) }	\\ 
\addlinespace[0.1cm]
Transformer  &    616   	&  217	 &  \cellcolor{aliceblue}0.73	&  \cellcolor{aliceblue}0.26   \\
AoANet   &         527	& 213	 &  \cellcolor{aliceblue}0.71	&  \cellcolor{aliceblue}0.28   \\
Vilbert  &  526     	&  161	 &  \cellcolor{aliceblue}0.76	&  \cellcolor{aliceblue}0.23   \\
OSCAR  &     630   	&  237	 &  \cellcolor{aliceblue}0.72	&  \cellcolor{aliceblue}0.27   \\
BLIP & 554 & 240 & \cellcolor{aliceblue}0.69   & \cellcolor{aliceblue}0.30  \\

TraCLIPS-Reward  &537 & 251 & \cellcolor{aliceblue}0.68   & \cellcolor{aliceblue}0.31 \\
BLIP-2 &498 &  239 & \cellcolor{aliceblue}0.67   & \cellcolor{aliceblue}0.32 \\
\bottomrule 

\end{tabular}
}
\vspace{-0.2cm}



\caption{Full amplified gender-to-object results. Comparison result between 1) initial bias via language model GPT-2; 2) Baseline BERT Mask Language Modeling (MLM) with \texttt{MASK} prediction without any visual context; 3) Object Gender Co-Occ; 4) our Gender Score estimation with \texttt{MASK} prediction on the karpathy test split.  \colorbox{aliceblue}{Our result} (bias ratio) is  \colorbox{blond}{similar to BERT} in some cases, as shown in Figure \ref{tb-fig:3} with \textit{man} instance with the \textit{pizza} bias. However, when the generated caption is more/less diverse, the MLM struggles with predicting the bias and correct gender. Unlike BERT, our score relies on additional parameters such as the similarity score to measure an accurate gender bias prediction.}




\label{tb:benh-additional}
\end{table}

 \section*{Baseline}


\noindent{\bf{Transformer \cite{Ashish:17}.}} We use a transformer model that is trained on bottom-up and top-down features  \cite{anderson2018bottom} with cross-entropy loss with learning rate $3e-04$.

\noindent{\bf{AoANet \cite{huang2019attention}}.} We utilize the model as out-of-the-box with the best-reported result by the authors. 

\noindent{\bf{Vilbert$_{\text{Large}}$ \cite{lu202012}}} (12-in-1: Multi-Task Vilbert) Since Vilbert is trained on 12 datasets, we also use the provided pre-trained model out-of-the-box extracted captions \cite{sabir2022belief}. 

\noindent{\bf{OSCAR$_{\text{Large}}$ \cite{li2020oscar}.}} To make a fair comparison with other pre-trained vision and language models, we fine-tuned the model on the COCO-Caption dataset with cross-entropy evaluation score without RL with a learning rate $5e-06$ (15 epochs), following the same paper setup.

\noindent{\bf{BLIP$_{\text{Large}}$ \cite{li2022blip}}.} We use the model with a beam search b=3 (best-reported result by the
authors with their provided checking points.

\noindent{\bf{TransCLIPS-Reward} \cite{cho2022fine}.} Transformer CLIPS-Reward best-proposed model by the authors \texttt{CLIP-S+Grammar} reward inherits biases from RL and thus generates caption w/o a specific gender (\eg  \textit{person}, \textit{guy}, etc.). Therefore, we adopt the combined CLIPS+CIDEr Rewards, which suffer less gender prediction error.

\noindent{\bf{BLIP-2 VIT-G OPT$_{\text{2.7B}} $\cite{li2023blip}}.}  We use the best image captioning model reported by the authors in the benchmark (beam search $b$=5) with Open Pre-trained Transformer (OPT)  Large Language Models (LLMs)  (with 2.7B parameters model), and scaled vision transformer 2B parameters model (with VIT-Giant 1843M parameters). 


\section*{Case Study}


\noindent{\bf{Gender Bias in Twitter.}} In this section, we  apply our proposed Gender Score to a subset of the Twitter user gender classification dataset\footnote{\url{https://www.kaggle.com/crowdflower/twitter-user-gender-classification}}, as shown in Table \ref{tb:twiter-gp}. We use a BERT-based keyword extractor\footnote{\url{https://github.com/MaartenGr/KeyBERT}} to extract the biased context from the sentence (\eg travel-man, woman-family), and we then employ the cloze probability\footnote{From cognitive linguistics, cloze probability is the probability of the target word completing that particular sentence.} \cite{gonzalez2007methods} to extract the probability of the context. We observe that some keywords have a strong  bias: women are associated  with keywords such as \textit{novel}, \textit{beauty}, and \textit{hometown}. Meanwhile, men are more  frequently related  to words such as \textit{gaming}, \textit{coffee}, and \textit{inspiration}.  Table \ref{tb:tw-result} shows a comparison result  between the frequency count method via ground truth and our score.

\begin{table}[ht!]\centering
\resizebox{0.8\columnwidth}{!}{
\begin{tabular}{lcc}
\toprule 
Model &   to-m &  to-w \\
\midrule 
\addlinespace[0.1cm]
\multicolumn{2}{l}{ Object Gender Co-Occ  \cite{zhao2017men}}	\\ 
\addlinespace[0.1cm]
Twitter Profile Data  &      0.55	&  0.44   \\
\hline 
\multicolumn{2}{l}{Gender Score  }	\\ 
\addlinespace[0.1cm]
\cellcolor{aliceblue}Twitter Profile Data  &    \cellcolor{aliceblue} 0.50  	&\cellcolor{aliceblue}  0.49	  \\

\bottomrule 

\end{tabular}
}
\vspace{-0.2cm}
\caption{Gender Ratio results on gender classification task via \colorbox{aliceblue}{Gender Score} and estimated ground truth.}


%
\label{tb:tw-result}
\end{table}

\begin{table}[t!]
\centering
\resizebox{0.7\columnwidth}{!}{
\begin{tabular}{lccc}

\hline 

Object &  + man  & + woman \\ 
\hline 
 


hoopskirt &    0.17 \DrawPercentageBar{0.17} &      \textbf{0.19}  \DrawPercentageBarNeg{0.19}  \\  
bobsled &   0.16 \DrawPercentageBar{0.17} &      \textbf{0.17}  \DrawPercentageBarNeg{0.17}  \\  
bagel  & 0.14 \DrawPercentageBar{0.17} &   0.14  \DrawPercentageBarNeg{0.14}  \\  
sliding door  & 0.14 \DrawPercentageBar{0.14} &   0.14  \DrawPercentageBarNeg{0.14}  \\  
 tobacco shop & 0.26 \DrawPercentageBar{0.26} &   \textbf{0.27}  \DrawPercentageBarNeg{0.27}  \\  
pizza  & \textbf{0.35} \DrawPercentageBar{0.35} &   0.32  \DrawPercentageBarNeg{0.32}  \\  
cellular telephone  & \textbf{0.55} \DrawPercentageBar{0.55} &  0.54  \DrawPercentageBarNeg{0.54}  \\  
refrigerator  & 0.17 \DrawPercentageBar{0.17} &   \textbf{0.23}  \DrawPercentageBarNeg{0.23}  \\  
washbasin  & 0.24 \DrawPercentageBar{0.24} &  0.24 \DrawPercentageBarNeg{0.24}  \\  
paddle  & \textbf{0.26} \DrawPercentageBar{0.26} &   0.24 \DrawPercentageBarNeg{0.24}  \\  
  ski  & \textbf{0.54} \DrawPercentageBar{0.54} &   0.45 \DrawPercentageBarNeg{0.45}  \\  
television  & \textbf{0.35} \DrawPercentageBar{0.35} &   0.30 \DrawPercentageBarNeg{0.30}  \\  
restaurant  &  0.28 \DrawPercentageBar{ 0.28} &   \textbf{0.37} \DrawPercentageBarNeg{0.37}  \\  
wine bottle  & 0.43 \DrawPercentageBar{0.43} &   \textbf{0.47} \DrawPercentageBarNeg{0.47}  \\  
soccer ball  & \textbf{0.57} \DrawPercentageBar{0.57} &   0.49 \DrawPercentageBarNeg{0.49}  \\  
overskirt & 0.11 \DrawPercentageBar{0.11} &  \textbf{0.16} \DrawPercentageBarNeg{0.16}  \\ 
ballplayer & \textbf{0.56} \DrawPercentageBar{0.56 } &   0.41 \DrawPercentageBarNeg{0.41}  \\  	
parking meter  & 	\textbf{0.25} \DrawPercentageBar{0.23} &   0.23 \DrawPercentageBarNeg{0.23}  \\          
umbrella & 0.28 \DrawPercentageBar{0.28} &  \textbf{0.30} \DrawPercentageBarNeg{0.30}  \\  
walking stick  & \textbf{0.13} \DrawPercentageBar{0.13} &   0.09 \DrawPercentageBarNeg{0.09}  \\ 	
bathing cap & 0.27 \DrawPercentageBar{0.27} &   \textbf{0.28} \DrawPercentageBarNeg{0.28}  \\  	
tape player  & \textbf{0.12} \DrawPercentageBar{0.12} &   0.09 \DrawPercentageBarNeg{0.09}\\          
whiskey jug  & \textbf{0.08} \DrawPercentageBar{0.08} &   0.07 \DrawPercentageBarNeg{0.07}  \\       
chocolate sauce  & 0.07 \DrawPercentageBar{0.07} &   \textbf{0.08} \DrawPercentageBarNeg{0.07}  \\  
dishwasher  & 0.13 \DrawPercentageBar{0.13} &  0.13 \DrawPercentageBarNeg{0.13}  \\ 
toyshop  & \textbf{0.17}\DrawPercentageBar{0.17} &  0.14 \DrawPercentageBarNeg{0.14}  \\ 
bassinet & 0.11  \DrawPercentageBar{0.11} &  \textbf{0.12} \DrawPercentageBarNeg{0.12}  \\ 
necklace & 0.19  \DrawPercentageBar{0.19} &  	\textbf{0.20} \DrawPercentageBarNeg{0.20}  \\ 
ipod  & 0.10 \DrawPercentageBar{0.10} &   \textbf{0.12} \DrawPercentageBarNeg{0.12}  \\  
dutch oven   & 0.24 \DrawPercentageBar{0.24 } &  \textbf{0.29} \DrawPercentageBarNeg{0.29 }  \\              
moving van & \textbf{0.32} \DrawPercentageBar{0.32} &   0.30 \DrawPercentageBarNeg{0.30}  \\  	
shower curtain  & 0.31 \DrawPercentageBar{0.31} &   \textbf{0.32} \DrawPercentageBarNeg{0.32}  \\  
park bench  & \textbf{0.21}  \DrawPercentageBar{0.21} &  0.20 \DrawPercentageBarNeg{0.20}  \\  
tennis ball   & 0.47  \DrawPercentageBar{0.47} &  \textbf{0.51}  \DrawPercentageBarNeg{0.51}  \\  
granny smith    & 0.18 \DrawPercentageBar{0.18} &  \textbf{0.22}  \DrawPercentageBarNeg{0.22}  \\  
passenger car     & 0.05 \DrawPercentageBar{0.05} &  \textbf{0.06}  \DrawPercentageBarNeg{0.06}  \\     
sweatshirt   & 0.20 \DrawPercentageBar{0.20} &  0.20  \DrawPercentageBarNeg{0.20}  \\   
reflex camera  & 0.05 \DrawPercentageBar{0.05} & \textbf{0.12} \DrawPercentageBarNeg{0.12}  \\ 
 banana   & 0.30 \DrawPercentageBar{0.30} &   \textbf{0.32} \DrawPercentageBarNeg{0.32}  \\ 
book jacket    & 0.28 \DrawPercentageBar{0.28} &   \textbf{0.31} \DrawPercentageBarNeg{0.31}  \\ 
ice cream   & 0.14 \DrawPercentageBar{0.14} &   \textbf{0.15} \DrawPercentageBarNeg{0.15}  \\ 
folding chair     & \textbf{0.25} \DrawPercentageBar{0.25} &   0.22 \DrawPercentageBarNeg{0.22}  \\ 
cheeseburger      & \textbf{0.22} \DrawPercentageBar{0.21} &  0.19 \DrawPercentageBarNeg{0.19}  \\ 
 	
 vending machine  & 0.29 \DrawPercentageBar{0.29 } &   	\textbf{0.33} \DrawPercentageBarNeg{0.33}  \\  
tricycle  & 0.28 \DrawPercentageBar{0.28} &   \textbf{0.31} \DrawPercentageBarNeg{0.31}  \\  




\hline 
\end{tabular}
}
\vspace{-0.2cm}
\caption{CLIP-based Gender Score on randomly selected objects from the test set. The baseline is a Transformer \cite{Ashish:17}  that is trained with Bottom-Up and Top-Down features \cite{anderson2018bottom}.}
\label{tb:gp}
\end{table}

\begin{figure}[t!]
\hspace*{-0.5in}
\centering 

\includegraphics[width=0.8\columnwidth]{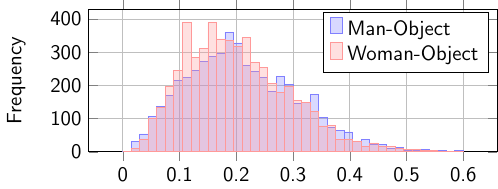}
\vspace{-0.2cm}
\caption{Probability distribution of our Gender Score \ie object-to-gender semantic bias score on the karpathy test split. As shown in the Figure, the distributions of object-to-gender $\in\{\text {man, woman}\}$ bias are similar in most instances, which indicate that not all object has a strong bias toward a specific gender.}
\label{fig:grpah}


\end{figure} 








\begin{table*}[t!]
\centering
\resizebox{\textwidth}{!}{
\begin{tabular}{lccc}

\hline 

Tweets  &   + man  & + woman & GT confidence \\ 
\hline 
 


BSc economics graduate &   0.00 \DrawPercentageBar{0} &      0.00 \DrawPercentageBarNeg{0} &  0.99 \DrawPercentageBar{0.99} \\  

Blood makes you related. Loyalty makes you family  & 0.17 \DrawPercentageBar{0} &      \textcolor{red}{0.27}  \DrawPercentageBarNeg{0.27} & 0.99 \DrawPercentageBar{0.99} \\

I'm the author of novels filled with family drama and romance.  &0.22 \DrawPercentageBar{0.22} &      \textcolor{red}{0.34}  \DrawPercentageBarNeg{0.34} & 0.99 \DrawPercentageBar{0.99} \\

 You are to me all that an angel could be.  &  0.12 \DrawPercentageBar{0.12} &     \textcolor{red}{0.13}  \DrawPercentageBarNeg{0.13} & 0.99 \DrawPercentageBar{0.99} \\

ask no questions and you'll get no lies & \textcolor{red}{0.15} \DrawPercentageBar{0.15}     &  0.13  \DrawPercentageBarNeg{0.13} & 0.99 \DrawPercentageBarNeg{0.99}  \\  

who cares what people think &  0.17 \DrawPercentageBar{0.17}   &    0.17  \DrawPercentageBarNeg{0.17} & 0.68 \DrawPercentageBarNeg{0.68}  \\

I know I'm mad because iv always been mad &  \textbf{0.10} \DrawPercentageBar{0.10} &   0.19  \DrawPercentageBarNeg{0.08} &   0.99 \DrawPercentageBar{0.99}\\

I'm floating!!!!! I read lots of them comic book things. &  \textbf{0.11}	 \DrawPercentageBar{0.11} &      0.10  \DrawPercentageBarNeg{0.10} &   0.99 \DrawPercentageBar{0.99}\\

weird in a mysterious way, i'm not a psycho & 0.12 \DrawPercentageBar{0.12} &      \textbf{0.13} \DrawPercentageBarNeg{0.13} & 0.99 \DrawPercentageBarNeg{0.99}  \\

there are two kinds of people in this world And I dont like them   & \textcolor{red}{0.28} \DrawPercentageBar{0.28} &      0.27  \DrawPercentageBarNeg{0.27} & 0.99 \DrawPercentageBarNeg{0.99}  \\

nothing more important to me than my family.   & 0.10 \DrawPercentageBar{0.10} &     \textbf{0.16}  \DrawPercentageBarNeg{0.16} & 0.99 \DrawPercentageBarNeg{0.99}  \\

do you believe in time travel?  & \textcolor{red}{0.28} \DrawPercentageBar{0.28} &      	0.24  \DrawPercentageBarNeg{0.24} & 0.99 \DrawPercentageBarNeg{0.99}  \\  

fan of the latest technology   &  0.05 \DrawPercentageBar{0.05} &     0.05  \DrawPercentageBarNeg{0.05} & 0.99  \DrawPercentageBar{0.99} \\  

life is not a dress rehearsal.  &  0.10 \DrawPercentageBar{0.10} &    \textbf{0.11}  \DrawPercentageBarNeg{0.11} & 0.99 \DrawPercentageBarNeg{0.99} \\

I love music and watching livestreams  & \textcolor{red}{0.13} \DrawPercentageBar{0.13} &     0.12  \DrawPercentageBarNeg{0.12} & 0.35
  \DrawPercentageBarNeg{0.35} \\

I don't get how people can get so anti-something. &   0.32 \DrawPercentageBar{0.32} &     \textbf{0.33}  \DrawPercentageBarNeg{0.33} & 0.99
  \DrawPercentageBarNeg{0.99} \\

I am me and that is who I want to be I wont change for anyone &   0.21 \DrawPercentageBar{0.21} &     \textbf{0.22} \DrawPercentageBarNeg{0.22} & 0.99
  \DrawPercentageBarNeg{0.99} \\

Left my hometown in search of myself.  &0.14 \DrawPercentageBar{0.14} &      \textcolor{red}{0.21}  \DrawPercentageBarNeg{0.21} &   0.99 \DrawPercentageBar{0.99}\\

My life is inspired by actual events & \textbf{0.20} \DrawPercentageBar{0.20} &      0.14  \DrawPercentageBarNeg{0.14} &   0.99 \DrawPercentageBar{0.99}\\

do whatever makes you happy
  &  0.25 \DrawPercentageBar{0.25} &     0.25  \DrawPercentageBarNeg{0.25} & 0.99  \DrawPercentageBarNeg{0.99} \\

 I write a beauty blog also love me some metal music !!!
 &  0.15 \DrawPercentageBar{0.15} &     	\textcolor{red}{0.17}  \DrawPercentageBarNeg{0.17 } & 0.99 \DrawPercentageBar{0.99} \\

sometimes I'm cynical. I have no filter. &   \textbf{0.13} \DrawPercentageBar{0.13} &     0.12  \DrawPercentageBarNeg{0.12} & 0.99
  \DrawPercentageBar{0.99} \\

I am a web and iOS developer   & 0.17 \DrawPercentageBar{0.17} &     \textcolor{red}{0.18}  \DrawPercentageBarNeg{0.18} & 0.99
  \DrawPercentageBar{0.99} \\

and if you like having secret little rendezvous &   0.13 \DrawPercentageBar{0.13} &     \textbf{0.18}  \DrawPercentageBarNeg{0.18 } & 0.99
  \DrawPercentageBarNeg{0.99} \\

 You can make great money by buying things cheaply
  & \textbf{0.04} \DrawPercentageBar{0.04} &      0.03  \DrawPercentageBarNeg{0.03} &   0.99 \DrawPercentageBar{0.99}\\

You may be a sinner but your innocence is mine.
   & \textbf{0.11} \DrawPercentageBar{0.11} &      0.10  \DrawPercentageBarNeg{0.10} &   0.99 \DrawPercentageBar{0.99}\\

I live off white wine and avocados &  0.24 \DrawPercentageBar{0.24} &      \textbf{0.25}  \DrawPercentageBarNeg{0.25} & 0.99 \DrawPercentageBarNeg{0.99}  \\

 be without fear. & 0.04 \DrawPercentageBar{0.04} &     0.19  \DrawPercentageBarNeg{0.04} & 0.99   \DrawPercentageBar{0.99} \\  

I love anything related to gaming.  &  \textbf{0.14} \DrawPercentageBar{ 0.14} &     0.13 \DrawPercentageBarNeg{0.13} & 0.99 \DrawPercentageBar{0.99} \\

There's such a lot of world to see & \textcolor{red}{0.28} \DrawPercentageBar{0.28} &     0.25 \DrawPercentageBarNeg{0.25} & 0.99 \DrawPercentageBarNeg{0.99} \\

I'm probably funnier than you. &  \textbf{0.34} \DrawPercentageBar{0.34} &     0.31  \DrawPercentageBarNeg{0.31} & 0.99
   \DrawPercentageBar{0.99} \\

I'm just here for sweet and detailed everythings     & 0.04 \DrawPercentageBar{0.04} &     0.04 \DrawPercentageBarNeg{0.04} & 0.99
  \DrawPercentageBarNeg{0.99} \\

You probably think this song is about you   & 0.16 \DrawPercentageBar{0.16} &      \textbf{0.20}  \DrawPercentageBarNeg{0.20} &   0.99
  \DrawPercentageBarNeg{0.99}\\

go after your dream &  0.01 \DrawPercentageBar{0.01} &     0.01  \DrawPercentageBarNeg{0.01} & 0.99
  \DrawPercentageBar{0.99} \\

Let's grab coffee &  \textbf{0.15} \DrawPercentageBar{0.15} &      0.12  \DrawPercentageBarNeg{0.12} &   0.99 \DrawPercentageBar{0.99}\\

We can dream a dream for us that no one else can touch
&  0.14 \DrawPercentageBar{0.14} &     \textcolor{red}{0.15} \DrawPercentageBarNeg{0.15} & 0.66 \DrawPercentageBar{0.66}  \\

 One day I'll be able to stop crying over people who aren't worth it
 &  0.21 \DrawPercentageBar{0.21} &     0.21  \DrawPercentageBarNeg{0.21} & 0.67  \DrawPercentageBarNeg{0.67} \\

Believe in something bigger than yourself and find your purpose.
 & \textcolor{red}{0.28} \DrawPercentageBar{0.28} &     0.25  \DrawPercentageBarNeg{0.25} & 0.99 \DrawPercentageBarNeg{0.99} \\


some say I'm a photographer who loves politics & 0.06 \DrawPercentageBar{0.06} &     0.07  \DrawPercentageBarNeg{0.07} & 0.99
  \DrawPercentageBar{0.99} \\

Student with an interest in politics&    0.19 \DrawPercentageBar{0.19} &     \textcolor{red}{0.20}  \DrawPercentageBarNeg{0.20} & 0.99
  \DrawPercentageBar{0.99} \\

 the world isn't as cruel as you take it to be  & \textbf{0.42} \DrawPercentageBar{0.42} &      0.37  \DrawPercentageBarNeg{0.37} &   0.99 \DrawPercentageBar{0.99}\\

It's hard to want to stay awake when everyone you meet
 &  \textbf{0.21} \DrawPercentageBar{0.21} &      0.19  \DrawPercentageBarNeg{0.19} &   0.99 \DrawPercentageBar{0.99}\\

You're such a hypocrite   & \textbf{0.36} \DrawPercentageBar{0.36} &      	0.30  \DrawPercentageBarNeg{0.30} & 0.32 \DrawPercentageBar{0.32}  \\

 Half human half dolphin and professional tea maker &  0.00 \DrawPercentageBar{0.00} &     0.00  \DrawPercentageBarNeg{0.00} & 0.99  \DrawPercentageBarNeg{0.99} \\

 you can now purchase my book word vomit in stores everywhere &  \textbf{0.03} \DrawPercentageBar{0.03} &     0.02  \DrawPercentageBarNeg{0.02} & 0.99 \DrawPercentageBar{0.99} \\

 i love watching ghost adventures   & 0.00 \DrawPercentageBar{0.00} &     0.00  \DrawPercentageBarNeg{0.00} & 0.36
  \DrawPercentageBarNeg{0.36} \\

im not a bad person but i did bad things  &\textcolor{red}{0.18} \DrawPercentageBar{0.18} &     0.14  \DrawPercentageBarNeg{0.14} & 0.99
  \DrawPercentageBarNeg{0.99} \\

 You have to understand that sometimes is hard to understand me. & \textbf{0.27} \DrawPercentageBar{0.27} &     0.25  \DrawPercentageBarNeg{0.25} & 0.99
  \DrawPercentageBar{0.99} \\

I mean Im in college or something like that
 & 0.20 \DrawPercentageBar{0.20} &      \textcolor{red}{0.21}  \DrawPercentageBarNeg{0.21} &   0.99 \DrawPercentageBar{0.99}\\

if u cant be happy u can at least be drunk & \textbf{0.17} \DrawPercentageBar{0.17} &      0.16  \DrawPercentageBarNeg{0.16} &   0.99 \DrawPercentageBar{0.99}\\

We are all addicted to something & 0.19 \DrawPercentageBar{0.19} &      \textbf{0.22}  \DrawPercentageBarNeg{0.22} & 0.99 \DrawPercentageBarNeg{0.99}  \\

It's a hard life in the mountains.   &  \textbf{0.11} \DrawPercentageBar{0.11} &     0.10  \DrawPercentageBarNeg{0.10} & 0.99  \DrawPercentageBar{0.99} \\

my heart is yet to be thawed  &  0.17 \DrawPercentageBar{0.17} &  \textbf{0.18} \DrawPercentageBarNeg{0.18} & 0.99 \DrawPercentageBarNeg{0.99} \\

I can't find my damn left sock   & 0.10  \DrawPercentageBar{0.10} &     \textbf{0.19}  \DrawPercentageBarNeg{0.19} &0.10
  \DrawPercentageBarNeg{0.35} \\

Only speak if what you have to say is more beautiful than the silence.  & \textcolor{red}{0.40} \DrawPercentageBar{0.40} &     0.39  \DrawPercentageBarNeg{0.39} & 0.99
  \DrawPercentageBarNeg{0.99} \\

left/right it doesn't matter where you go. important is u choose a path and didn't    & \textbf{0.19} \DrawPercentageBar{0.19} &     0.18  \DrawPercentageBarNeg{0.18} & 0.99
  \DrawPercentageBar{0.99} \\

I'm a Christian. I love Jesus and pray to always glorify Him.  & \textcolor{red}{0.23} \DrawPercentageBar{0.23} &      	0.18 \DrawPercentageBarNeg{0.18} &   0.99 \DrawPercentageBarNeg{0.99}\\

Don't have a lot to say & \textbf{0.30} \DrawPercentageBar{0.30} &      0.25 \DrawPercentageBarNeg{0.25} &   0.99 \DrawPercentageBar{0.99}\\

\hline 

\end{tabular}
}
\vspace{-0.2cm}

\caption{Gender Score on user gender classification task (Twitter). We observe that some keywords have a strong  bias: women are associated  with keywords such as \textit{novel}, \textit{beauty}, and \textit{hometown}. Meanwhile, men are more  frequently related  to words such as \textit{gaming}, \textit{coffee}, and \textit{inspiration}. Highlighted in red color, the Gender Score disagrees with human estimation confidence due to gender-context bias.}
\label{tb:twiter-gp}
\end{table*}

\end{document}